\documentclass[10pt,journal,compsoc]{IEEEtran}
\ifCLASSOPTIONcompsoc
  \usepackage[nocompress]{cite}
\else
  \usepackage{cite}
\fi
\ifCLASSINFOpdf
\else
\fi

\usepackage{epsfig}
\usepackage{graphicx}
\usepackage{amsmath}
\usepackage{amssymb}
\usepackage{dirtree}
\usepackage{subcaption}
\usepackage{booktabs}
\begin{document}
%
\title{Small Data Challenges in Big Data Era: A Survey of Recent Progress on Unsupervised and Semi-Supervised Methods}
%
%
%
%

\author{Guo-Jun~Qi,~\IEEEmembership{Senior Member,~IEEE,}
        and~Jiebo~Luo,~\IEEEmembership{Fellow,~IEEE,}
\IEEEcompsocitemizethanks{\IEEEcompsocthanksitem G.-J. was with the Laboratory for MAchine Perception and LEarning (maple-lab.net) and the Futurewei Technologies, Bellevue,
WA, 98004.\protect\\
E-mail: guojunq@gmail.com
\IEEEcompsocthanksitem J. Luo was with the Department of Computer Science, University of Rochester, Rochester, NY 14627.\protect\\
E-mail: jluo@cs.rochester.edu}
}

%
%

\markboth{Qi \MakeLowercase{\textit{et al.}}: Small Data Challenges in Big Data Era: A Survey of Recent Progress on Unsupervised and Semi-Supervised Methods}%
{Qi \MakeLowercase{\textit{et al.}}: Small Data Challenges in Big Data Era: A Survey of Recent Progress on Unsupervised and Semi-Supervised Methods}
%



\IEEEtitleabstractindextext{%
\begin{abstract}
 Representation learning with small labeled data have emerged in many problems, since the success of deep neural networks often relies on the availability of a huge amount of labeled data that is expensive to collect. To address it, many efforts have been made on training sophisticated models with few labeled data in an unsupervised and semi-supervised fashion. In this paper, we will review the recent progresses on these two major categories of methods. A wide spectrum of models will be categorized in a big picture, where we will show how they interplay with each other to motivate explorations of new ideas. We will review the principles of learning the transformation equivariant, disentangled, self-supervised and semi-supervised representations, all of which underpin the foundation of recent progresses. Many implementations of unsupervised and semi-supervised generative models have been developed on the basis of these criteria, greatly expanding the territory of existing autoencoders, generative adversarial nets (GANs) and other deep networks by exploring the distribution of unlabeled data for more powerful representations. We will discuss emerging topics by revealing the intrinsic connections between unsupervised and semi-supervised learning, and propose in future directions to bridge the algorithmic and theoretical gap between transformation equivariance for unsupervised learning and supervised invariance for supervised learning, and unify unsupervised pretraining and supervised finetuning. We will also provide a broader outlook of future directions to unify transformation and instance equivariances for representation learning, connect unsupervised and semi-supervised augmentations, and explore the role of the self-supervised regularization for many learning problems.
\end{abstract}

\begin{IEEEkeywords}
Unsupervised methods, semi-supervised methods, domain adaptation, transformation equivariance and invariance, disentangled representations, generative models, auto-encoders, generative adversarial networks, auto-regressive models, flow-based generative models, transformers, self-supervised methods, teach-student models, instance discrimination and equivariance.
\end{IEEEkeywords}}

\maketitle

\IEEEdisplaynontitleabstractindextext

%
\IEEEpeerreviewmaketitle

\IEEEraisesectionheading{\section{Introduction}\label{sec:intro}}

\begin{figure*}[t]
    \centering
    \begin{subfigure}[c]{0.7\textwidth}
        \includegraphics[width=\textwidth]{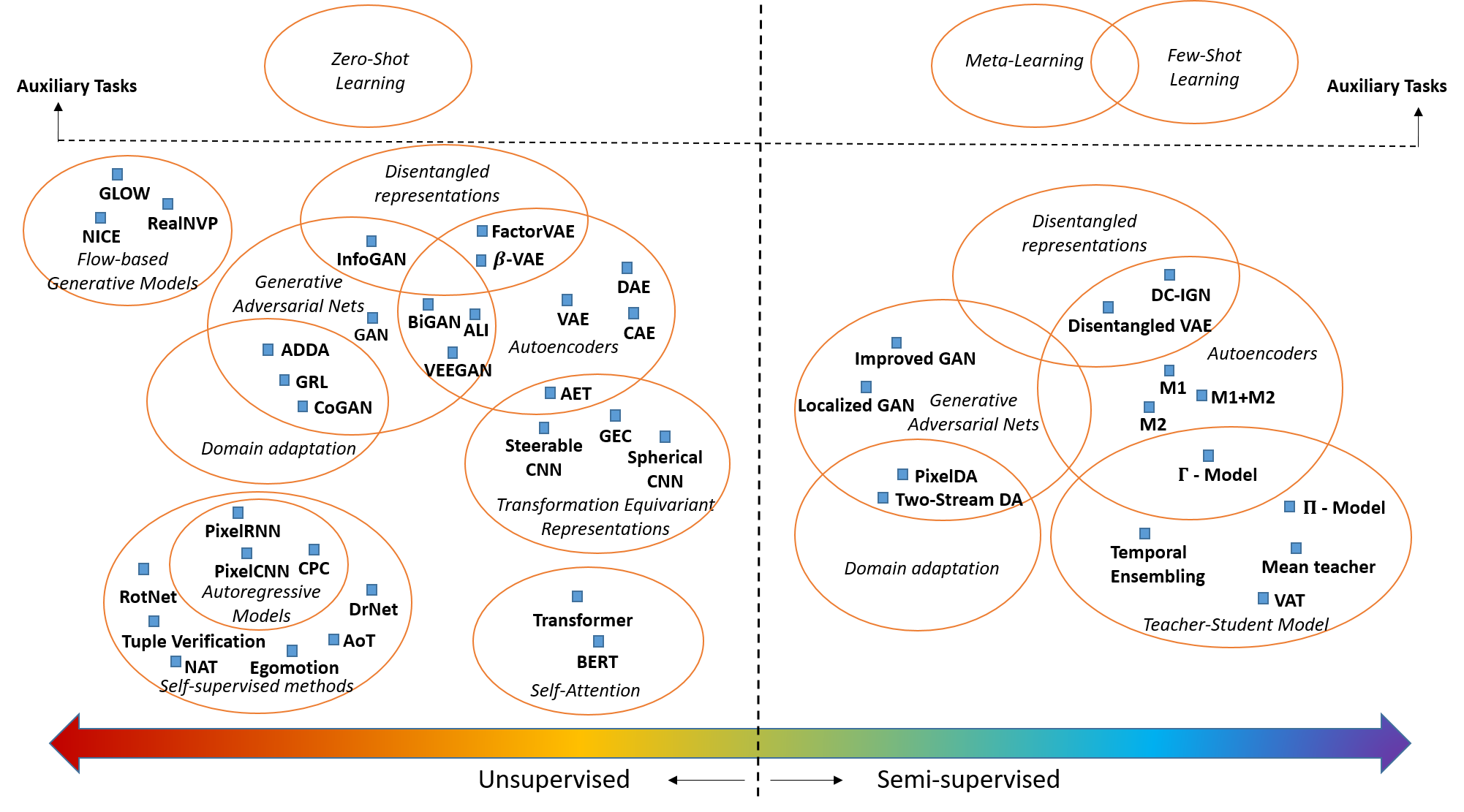}
    \end{subfigure}\\
    ~ 
    \caption{An overview of the landscape of unsupervised and semi-supervised methods. This figure shows the relations between different methods, and where they intersect with each other. Please refer to Figure~\ref{fig:small_data} for the categorization of these methods for this survey.}\label{fig:landscape}
\end{figure*}

\IEEEPARstart{T}{his} paper aims at a comprehensive survey of recent progresses on unsupervised and semi-supervised methods addressing the challenges of training models with a small number or none of labeled data when a large volume of unlabeled data are available. The success of deep learning often hinges on the availability of a large number of labeled data, where millions of images are labeled to train the deep neural networks \cite{krizhevsky2012imagenet,he2016deep} to enable these models to be on par with or even surpass the human performances.

However, in many cases, it is challenging to collect a sufficiently large number of labeled data, and this inspires many research efforts on exploring the unsupervised information beyond labeled data to train robust models for various learning tasks.

\begin{itemize}
  \item {\bf Unlabeled data.} While the number of labeled data would be extremely {\em small}, unlabeled data could be remarkably {\em big}.  The distribution of those unlabeled data provides important clues on learning robust representations that are generalizable to new learning tasks. The unlabeled data can be leveraged in both an unsupervised and a semi-supervised fashion, depending on whether additional labeled examples are leveraged to train models. Unlabeled data can also assist models to close the domain gap between different tasks, and this leads to a large category of unsupervised and semi-supervised domain adaptation approaches.
  \item {\bf Auxiliary tasks.} Auxiliary tasks can also be leveraged to mitigate small data problems as an important source of side information. For example, a related task can be a learning problem on a disjoint set of concepts that are related to the target task. This falls into the category of Zero-Shot Learning (ZSL) and Few-Shot Learning (FSL) problems. In a generalized sense, the ZSL problem can be viewed as an unsupervised learning problem with no labeled example on the target task, while the FSL is semi-supervised with few available labeled data. Both aim to transfer the semantic knowledge or the knowledge of learning (e.g., meta-learning \cite{finn2017model,jamal2018task,li2019learning}) from the source tasks to the target ones.
\end{itemize}

The focus of this survey is on the unsupervised and semi-supervised methods by exploring the unlabeled examples to address the small data problem. Although we will not review the ZSL and FSL methods that leverage the information from auxiliary tasks, it would be beneficial for us to start by looking at all these methods in a big picture. This will give us a better understanding of where we are in  the journey towards conquering the small data challenges.

Different ways of leveraging various sources of information lead to a wide spectrum of learning methods to address the  challenge of few labeled data from different perspectives as illustrated in Figure~\ref{fig:landscape}. In the appendix, we also provide a chart of unsupervised and semi-supervised learning in Section~\ref{fig:small_data}.

\subsection{Unsupervised Methods}
At the leftmost end of the spectrum are unsupervised methods trained without labeled data.
These unsupervised methods seek to learn representations that are sufficiently generalizable to adapt to various learning tasks in future. In this case, the representations learned from unsupervised methods are usually assessed based on the performances of downstream classification tasks on top of these representations.

A variety of principles and models have been devoted to training unsupervised representations. As shown in Figure~\ref{fig:small_data}, we will review them from several different perspectives. First, we will review the emerging principle of {\em Transformation Equivariant Representations}(TER) pioneered in Hinton's seminal work \cite{hinton2011transforming} as well as the recent formulation of unsupervised training of such representations \cite{zhang2019aet}. It follows by reviewing a number of generative networks representative in many recent models, including the variants of auto-encoders, Generative Adversarial Nets (GANs), Flow-based Generative Networks, and Transformers (See Figure~\ref{fig:small_data}). The principle of learning {\em disentangling representations} from these generative models is also central to many unsupervised methods, and we will review them on how to extract interpretable generative factors from unlabeled data. Finally, self-supervised methods constitutes a large category of unsupervised models, and we will review autoregressive models as well as the self-supervised training of image and video representations.

Zero-Shot Learning (ZSL) also sits on the left end of the spectrum by mining the auxiliary tasks usually on a disjoint set of concepts.  Compared with the pure unsupervised methods, it often explores the semantic correlations between concepts by word embedding and visual attributes, and uses them to transfer the knowledge from the source to the target concepts. Given a new sample, the zero-shot learning can assign it to an unseen concept with its semantic embedding closest to the representation of the sample. The ZSL is unsupervised because the training examples are not labeled on the unseen concepts that the ZSL aims to learn. We refer the interested readers to a more detailed review of ZSL methods \cite{fu2017recent}.


\subsection{Semi-Supervised Methods}
Along the spectrum to the right are the semi-supervised methods, which explore both unlabeled and labeled examples to train the models. The idea lies in that unlabeled examples provides important clues on how data are generally distributed in the space, and a robust model can be trained by exploring this distribution. For example, a robust model ought to make stable and smooth predictions under random transformations (e.g., translations, rotations, flipping or even random perturbations by a GAN \cite{qi2018global}) along the direction of data manifold, or avoid from placing its decision boundary on high density areas of data distribution.

Along this direction, as shown in Figure~\ref{fig:small_data}, we will review semi-supervised generative models extending their unsupervised counterparts, such as semi-supervised auto-encoders and GANs, as well as their disentangled representations. A variety of teach-student models will also be reviewed by encouraging the consistency between the teacher and student models on both labeled and unlabeled data to train semi-supervised models. They can be categorized by different ways of the teacher models being obtained -- by either applying random and adversarial perturbations \cite{laine2016temporal,miyato2018virtual} to or  averaging over an ensemble of student models \cite{laine2016temporal}.

In the spectrum of semi-supervised methods also resides Few-Shot Learning (FSL) when auxiliary tasks on a disjoint set of concepts are leveraged to improve the model training. On one hand, it is like zero-shot learning when conceptual correlations can be used to share information between different concepts through their embedded representations. On the other hand, a group of auxiliary tasks can be sampled from a collection of base concepts, and a meta-model can be trained to distill the knowledge of how to update models with few examples (e.g., the initial point, and the rule of updating model parameters) \cite{jamal2018task} along with the unlabeled examples in a semi-supervised fashion \cite{li2019learning,ren2018meta,ma2019affinitynet}. Thus, the FSL can be viewed as a semi-supervised problem that has few labeled examples available on the target concepts, along with many examples labeled on auxiliary concepts (which thus should be viewed as unlabeled on the target concepts). For a comprehensive review of FSL, the interested readers can refer to \cite{chen2018a}.



\subsection{Connections between Unsupervised and Semi-Supervised Learning}
We will show that existing unsupervised and semi-supervised methods share many common ideas and principles.
One of core ideas in both methods lies in exploring the crucial role of unlabeled data and the distributions in unsupervised and semi-supervised training of representations, no matter if labeled data are involved. For example, both unlabeled and labeled data can be augmented under various forms of transformations and noises to explore their invariance and equivariance. Such data augmentations underly many unsupervised and semi-supervised methods to regularize the model training \cite{wu2018unsupervised,zhang2019aet,qi2019avt,rasmus2015semi,laine2016temporal,tarvainen2017mean,berthelot2019mixmatch,sohn2020fixmatch} or find the model vulnerability to make it more robust \cite{kurakin2016adversarial,szegedy2013intriguing,miyato2018virtual}.
Many unsupervised models such as Auto-Encoders, GANs and disentangled representations have also been tailored into semi-supervised counterparts by conditioning on both data and labels.


Indeed, unsupervised and semi-supervised methods share many common principles that are surprisingly insightful and important, which deserves our attentions in future directions. We will sort out these principles in Section~\ref{sec:related} and outline its future directions. In particular, we will discuss emerging topics to bridge the algorithmic and theoretical gap between {\em transformation equivariance} for unsupervised learning and {\em transformation invariance} for supervised learning, and combine {\em unsupervised pretraining} and {\em supervised finetuning}. We will also provide an outlook of future directions to unify {\em transformation and instance equivariances} for representation learning, connect {\em unsupervised and semi-supervised augmentations}, and explore the role of the {\em self-supervised regularization} for various learning problems.
We expect a more general learning theory and framework can be developed to reveal the connections between unsupervised and semi-supervised learning.

The remainder of this paper is organized as follows. Unsupervised methods will be reviewed in Section~\ref{sec:unsupervised}, followed by a survey of semi-supervised methods in Section~\ref{sec:ssm}.
We will also review unsupervised and semi-supervised domain adaptations in Section~\ref{sec:domain}.
After reviewing existing works, we will elaborate on emerging topics and future directions in Section~\ref{sec:related} that will connect unsupervised and semi-supervised learning in multiple future directions.
Finally, we will conclude the survey in Section~\ref{sec:concl}.

\section{Unsupervised Methods}\label{sec:unsupervised}
In this section, we will survey the literature on learning unsupervised representations. The goal of training an unsupervised representation  from unlabeled examples is to ensure it can generalize to new tasks in future.

We will start the review with the emerging principle of learning Transformation Equivariant Representations, to a variety of representative generative models and their disentangled representations of interpretable generative factors, and to various self-supervised methods for training image and video representations.



\subsection{Unsupervised Representation Learning}

The methods for training unsupervised representations roughly fall into the following three groups of research.
\begin{itemize}
\item {\noindent\bf Transformation-Equivariant Representations.} Recently, learning transformation-equivariant representations (TERs) from unlabeled data has attracted many attentions  in both unsupervised and supervised methods. In particular, a good TER equivaries with different types of transformations so that the scene structure in an image can be compactly encoded into its representation. Then the successive problems for recognizing unseen visual concepts can be performed on top of the trained TER. The notion of TER was originally proposed by Hinton et al. \cite{hinton2011transforming} in introducing capsule nets and it has been formalized in various ways. We will review it in Section~\ref{sec:ter}.
\item {\noindent \bf Generative Models.} Auto-Encoders, Generative Adversarial Nets and many other generative models have been widely studied in unsupervised learning problems, from which compact representations can be learned to characterize the generative process for unlabeled data. We will review the learning and inference problems for these models, as well as discuss the disentanglement of the resultant representations into generative factors that can interpret both intrinsic and extrinsic data variations. More generative models besides the auto-encoders and GANs will also be reviewed in Section~\ref{sec:gan}.
\item {\noindent\bf Self-Supervised Methods.} There also exist a large variety of self-supervisory signals to train models without access to any labeled data, including auto-regressive models that are self-supervised to reconstruct data themselves. We will review different genres of self-supervisory signals for learning unsupervised representations in Section~\ref{sec:self}.
\end{itemize}
We also evaluate these unsupervised methods in Section~\ref{sec:ul_eval} in the appendix.

\subsection{Transformation-Equivariant Representations}\label{sec:ter}
Before we start to review the methods of unsupervised representation learning, it is beneficial to ponder over what properties 
ought to be possessed by a good representation, particulary from the great success of the Convolutional Neural Networks (CNNs). This should lay the foundation for the practices in learning unsupervised representations.

Although a solid theory is still lacking, it is thought that both {\em equivalence} and {\em invariance} to image translations play a critical role in the success of CNNs, particularly for supervised classification tasks \cite{hinton2011transforming,krizhevsky2012imagenet}. A typical Convolutional Neural Network (CNN) consists of two parts: the {\em feature maps} of input images through multiple convolutional layers, and the classifier of {\em fully connected layers} mapping the feature maps to the target labels.

While the resultant feature maps are equivariant to the translation of an input image, the fully connected classifier should be predict labels invariant to any transformations. Before the concept of learning Transformation-Equivariance Representations (TER) was proposed by Hinton et al. \cite{cohen2016group,cohen2018intertwiners,sabour2017dynamic,hinton2011transforming}, most attentions have been paid on the transformation invariance criterion to train supervised models by minimizing the classification errors on labeled images augmented with various transformations \cite{krizhevsky2012imagenet}. Unfortunately, it is impossible to directly apply transformation invariance to train an unsupervised representation -- without the guidance of label supervision, this would lead to a trivial representation invariant to all examples.

Thus, it is a natural choice to adopt the transformation equivariance as the criterion to train an unsupervised representation, hoping it could be generalizable to unseen tasks without knowledge their labels. This is contrary to the criterion of transformation invariance that tends to tailor the learned representations more specialized to the labels of given tasks.
Indeed, it is straightforward to see that the feature maps generated through convolutional layers equivary with the translations -- the feature maps of translated images are also shifted in the same way subject to edge padding effect \cite{krizhevsky2012imagenet}. This inspires many works to generalize this idea to consider more types of transformations beyond translations (e.g., general image warping and projective transformations) \cite{cohen2016group}. This can learn a good representation of images by encoding their intrinsic visual structures that equivary with many transformations.


Along this line of research, Group-Equivariant Convolutions (GEC) \cite{cohen2016group} have been proposed by directly training feature maps as a function of different transformation groups. The resultant feature maps are proved to equivary exactly with designated transformations. However, the form of group-equivariant convolutions is strictly defined, which limits the flexibility of its representation in many applications. Alternatively, a more flexible way to enforce transformation equivariance is explored by maximize the dependency between the resultant representations and the chosen transformations, which results in Auto-Encoding Transformation (AET) \cite{zhang2019aet}. Compared with GEC, the AET does not exactly comply with the criterion of transformation equivariance, in pursuit for the flexibility in the form of unsupervised representations.



\subsubsection{Group-Equivariant Convolutions}
Consider a group $\mathcal G$, which could consist of compositions of various transformations such as rotations, translations, and mirror reflections. The goal of Group-Equivariant Convolution (GEC) is to produce feature maps that equivary to all transformations $ g\in \mathcal G$ from the group.

To formally introduce the concept of transformation equivariance, we can view an input image and a feature map $f$ as a function over an image grid $\mathbb Z^2$,
$$f: \mathbb Z^2 \rightarrow \mathbb R,$$
where $f(p)$ gives the feature at a pixel location $p$. For simplicity, we only consider a single channel feature map, but it can be directly extended to multi-channel scenario without any difficulty.

When a transformation $g\in \mathcal G$ is applied to $f$, it results in a transformed image or feature map $L_gf$:
$[L_gf](x) = [f\circ g^{-1}] = f(g^{-1} x)$. Then we say a convolution with a kernel filter $\psi$ is transformation equivariant to $g$, if $[L_g f] \star \psi = L_g [f\star\psi]$, that is the convolution with a transformed input equals to the transformation of the convolution with the original input.

To enable the transformation equivariance, in GEC, a feature map is considered as a function of the group $\mathcal G$, that is defined as $f:\mathcal G \rightarrow \mathbb R$.

Then the group convolution with an input image $f$ on $\mathbb Z^2$ is defined as $
[f\star\psi](g) = \sum_{y\in \mathbb Z^2} f(y)\psi(g^{-1} y)$,
yielding a group convolved feature map $[f\star\psi]$ defined over $\mathcal G$. Thus, all the feature maps after the input image are functions of $\mathcal G$, and the group convolution of such a feature map $f$ with a filter $\psi$ is defined as $[f\star\psi](g) = \sum_{h\in \mathcal G} f(h)\psi(g^{-1}h)$, where the filter $\psi$ is also defined on $\mathcal G$. If we restrict the group $\mathcal G$ to translation, it is not hard to show that the group convolution reduces to a conventional convolution.

Cohen and Welling \cite{cohen2016group} have proved the transformation equivariance of the above group convolutions,
\[
\begin{aligned}
\left[[L_u f] \star \psi\right](g) 
= \left[L_u[f\star \psi]\right](g)
\end{aligned}
\]
where $[L_u f](h) = f(u^{-1}h)$ defines the operator $L_u$ of applying a transformation $u$ to the input $f$. This shows the convolution of a transformed input is equal to the transformation of the convolved input, i.e., the {\em transformation equivariance}.

The group convolutions are often trained in a supervised fashion to represent images together with some classification layers (e.g., fully connected and softmax layers) in a neural network \cite{cohen2016group}. In principle, unsupervised training of them can also be performed by treating them as the encoder in an auto-encoder architecture. Moreover, there exists an efficient implementation by decomposing group convolutions into a filter transformation and a planar convolution \cite{cohen2016group}.

The idea of training group-equivariant representations has been extended to explore the transformation equivariance in more scenarios. For example, the group equivariant capsule nets combine the group-equivariant convolutions with the dynamic routing mechanism to train capsule nets \cite{lenssen2018group}; Spherical images are analyzed in the SO(3) group by Spherical CNNs \cite{cohen2018spherical}, while the equivariance properties of steerable representations have be studied in the SO(2) group by Steerable CNNs \cite{cohen2016steerable}. For more implementation details, we refer the readers to \cite{lenssen2018group,cohen2016steerable,cohen2016group}.


\subsubsection{Auto-Encoding Transformations}
Although group convolutions guarantee the transformation equivariance mathematically, they have a much restricted form of feature maps as a function of the considered transformation group. In many applications, we often prefer more flexible forms of representations that can be trained in an {\em unsupervised} fashion by exploring the distribution of unlabeled data. In this section, we will review the recently proposed paradigms of Auto-Encoding Transformations (AET) \cite{zhang2019aet} as well as the variational approach Auto-encoding Variational Transformations (AVT) \cite{qi2019avt}.

\vspace{2mm}
{\noindent\bf Auto-Encoding Transformations}
\vspace{2mm}

Unlike the conventional Auto-Encoding Data (AED) paradigm that learns representations by reconstructing data, the AET seeks to train the unsupervised model by decoding transformations from the representations of original and transformed images.
It assumes that if a transformation can be reconstructed, the representations should contain all necessary information about the visual structures of images before and after the transformation such that the representations are transformation equivariant. Moreover, there is no restriction on the form of the representations, and this makes it flexible to choose a suitable form of representations for future tasks.

Formally, consider a transformation $\mathbf t$ sampled from a distribution $p(\mathbf t)$, along with an image $\mathbf x$ drawn from a data distribution $p(\mathbf x)$. By applying $\mathbf t$ to $\mathbf x$, one transforms $\mathbf x$ to $\mathbf t(\mathbf x)$. Then the AET aims at learning an encoder $E_\theta: \mathbf x \mapsto E_\theta(\mathbf x)$ with the parameters $\theta$, which extracts the representation $E_\theta(\mathbf x)$ of the given sample $\mathbf x$.
Meanwhile, a transformation decoder $D_\phi: \left[E_\theta(\mathbf x), E_\theta(\mathbf t(\mathbf x))\right]\mapsto \hat {\mathbf t}$ is also learned, which estimates $\hat{\mathbf t}$ of the input transformation $\mathbf t$ by decoding it from the representations of original and transformed images.


The learning problem of Auto-Encoding Transformations (AET) boils down to learn the representation encoder $E_\theta$ and the transformation decoder $D_\phi$ jointly. For this purpose, the AET can be trained by minimizing the following reconstruction error $\ell(\mathbf t, \hat{\mathbf t})$  between a transformation $\mathbf t$ and its estimate $\hat{\mathbf t}$,
$$
\min_{\theta,\phi} \mathop\mathbb E\limits_{\mathbf t \sim p(\mathbf t), \mathbf x\sim p(\mathbf x)}\ell(\mathbf t, \hat{\mathbf t})
$$
where the estimate $\hat{\mathbf t}$ of the transformation is a function of the encoder $E_\theta$ and the decoder $D_\phi$ such that $\hat{\mathbf t}= D_\phi\left[E_\theta(\mathbf x), E_\theta(\mathbf t(\mathbf x))\right]$,
and the expectation $\mathbb E$ is taken over the sampled transformations and images.
Then, the network parameters of $E_\theta$ and $D_\phi$ are jointly updated over mini-batches by back-propagating the gradient of the loss $\ell$.

In \cite{zhang2019aet}, three types of transformations have been considered in the AET model: parametric transformations, GAN-induced transformations and non-parametric transformations. This shows a wide spectrum of transformations can be integrated into the AET model.

\vspace{2mm}
{\noindent\bf Autoencoding Variational Transformation}
\vspace{2mm}

From an information-theoretic point of view, Qi et al.~\cite{qi2019avt} propose an alternative Auto-encoding Variational Transformation (AVT) model that reveals the connection between the transformations and representations by maximizing their mutual information.  It assumes that a good TER ought to maximize its probabilistic dependency on transformations, such that the representation contains the intrinsic information to decode the transformations when the visual structures of images are transformed extrinsically.

Formally, the representation $\mathbf z$ of a transformed image $\mathbf t(\mathbf x)$ is specified by the mean $f_\theta(\mathbf t(\mathbf x))$ and the variance $\sigma_\theta (\mathbf t(\mathbf x))$ in the AVT, such that
$$
\mathbf z = f_\theta(\mathbf t(\mathbf x)) + \sigma_\theta (\mathbf t(\mathbf x)) \circ \epsilon
$$
where $\epsilon$ is drawn from a normal distribution $\mathcal N(\mathbf 0, \mathbf I)$, $\circ$ denotes the element-wise product and $\theta$ is the model parameters.

With this probabilistic representation, the mutual information can be maximized to learn $\theta$, that is
\begin{equation}\label{eq:avt}
\max_\theta \mathbb E_{\mathbf x \sim p(\mathbf x)} I(\mathbf t, \mathbf z|\mathbf x).
\end{equation}

Directly maximizing the mutual information could be intractable and a variational lower bound
$$
I(\mathbf t, \mathbf z|\mathbf x) \geq \mathbb E_{p_\theta (\mathbf t, \mathbf z|\mathbf x)} q_\phi (\mathbf t|\mathbf z,\mathbf x).
$$
has been derived by introducing a surrogate transformation decoder $q_\phi(\mathbf t|\mathbf z, \mathbf x)$ that is the conditional probability of the transformation $\mathbf t$ on the representation $\mathbf z$ and the image $\mathbf x$.

This enables us to jointly train the representation encoder $p_\theta$ and the transformation decoder $q_\phi$ efficiently by maximizing the above lower bound of the mutual information. We refer the interested readers to \cite{qi2019avt} for more details.

\subsection{Generative Representations}\label{sec:gan}
Generative models, such as Generative Adversarial Nets \cite{goodfellow2014generative}, Auto-Encoders and their variants have emerged as powerful tools to extract expressive representations from unlabeled data in an unsupervised fashion. In this subsection, we will review several directions of representation learning based on the unsupervised models, particularly GANs and auto-encoders as well as their representation disentangling counterparts for modeling independent and interpretable generative factors that are useful for many downstream tasks.

We will show that these generative models are largely related. For example, GANs rely on learning an encoder to infer the representation from data \cite{donahue2016adversarial,dumoulin2016adversarially} and reduce mode collapse \cite{srivastava2017veegan}, while the auto-encoders can be enhanced with the adversarial training to generate sharper reconstruction of data \cite{larsen2015autoencoding} from the whole space of latent codes \cite{makhzani2015adversarial}. Various forms of disentangled representations are also learned based on these generative models, opening an active research direction towards extracting, disentangling and interpreting generative factors from representations.


\subsubsection{Auto-Encoders}


Auto-Encoders and many variants \cite{rifai2011contractive,kingma2013auto,vincent2008extracting,vincent2010stacked} are the generative models seeking to reconstruct the input data by jointly training a pair of encoder (inference component) and decoder (reconstructor component). Here we will review the Variational Auto-Encoders (VAE) \cite{kingma2013auto} as well as the Denoising Auto-Encoders (DAE) \cite{vincent2008extracting,vincent2010stacked} and Contractive Auto-Encoders (CAE) \cite{rifai2011contractive}, which are closely related with the regularization mechanisms for disentangled representations in Section~\ref{sec:dr} and semi-supervised methods in Section~\ref{sec:tsm}.

\vspace{2mm}
{\noindent \bf Variational Auto-Encoders}
\vspace{2mm}

The Variational Auto-Encoder (VAE) \cite{kingma2013auto} trains an auto-encoder model by maximizing the variational lower bound of the marginal data likelihood $p_\theta(\mathbf x)$ of a parameterized model $p_\theta$. For this, a variational encoder $q_\phi(\mathbf z|\mathbf x)$ is used to approximate the intractable posterior $p_\theta (\mathbf z|\mathbf x)$, resulting in the following inequality to lower bound the marginal likelihood:
$$
\log p_\theta (\mathbf x) \geq \mathbb E_{q_\phi(\mathbf z|\mathbf x)} [\log p_\theta(\mathbf x|\mathbf z)] - D_{KL}\left(q_\phi(\mathbf z|\mathbf x)|| p(\mathbf z)\right)
$$
where $p(\mathbf z)$ is the prior of representation, and $p_\theta(\mathbf x|\mathbf z)$ is the decoder.  Reparameterization trick is also introduced to sample from $q_\phi(\mathbf z|\mathbf x)$ as
$$
\mathbf z = g_\phi(\mathbf x, \boldsymbol\epsilon) = \boldsymbol \mu_\phi(\mathbf x) + \boldsymbol \sigma_\phi(\mathbf x) \odot \boldsymbol\epsilon
$$
where $\boldsymbol\epsilon$ is randomly drawn from a simple Gaussian distribution with zero mean and unit deviation, and $\odot$ is the element-wise product. In this way, the model parameters $\phi$ are separated from the random noises, and thus the error signals can be back-propagated through the neural network to train the VAE.

Later on, when reviewing the disentangled representations in Section~\ref{sec:dr}, we will see that the VAE provides a powerful tool to study and implement the representation disentanglement to provide interpretable generative factors.

\vspace{2mm}
{\noindent \bf Towards Robust Auto-Encoders}
\vspace{2mm}

Both Denoising Auto-Encoders (DAE) \cite{vincent2008extracting} and Contractive Auto-Encoders (CAE) \cite{rifai2011contractive} aim to learn robust representations insensitive to noises on input data.

Unlike the typical auto-encoders, the DAE \cite{vincent2008extracting} takes noise-corrupted samples as input and attempts to reconstruct original data. This forces the neural networks to learn the robust representations that can be used to recover the uncorrupted clean data. There are many ways to corrupt data. For example, some parts of input data can be randomly removed and the DAE attempts to recover the missing parts; an image can also be randomly transformed by rotations, translations and mirror flips, and the DAE aims to learn robust representations from which the original image before the transformation can be recovered.

The CAE \cite{rifai2011contractive} learns the robust representations in a different way. Rather than relying on a decoder to reconstruct the original data in the DAE, the CAE directly penalizes the changes of representations learned by the encoder $E$ in presence of the small perturbations on input data. This results in the following penalty on the Frobenius norm of Jacobi matrix around an input sample $\mathbf x$ to train the CAE
$$
\|J_E(\mathbf x)\|_F^2 = \sum_{i,j} \left(\dfrac{\partial E_i (\mathbf x)}{\partial \mathbf x_j}\right)^2
$$
where $E_i$ denotes the $i$th element of the encoded representation of $\mathbf x$.

The idea of regularizing the model training by adding noises to the model input or even model itself has led to many regularization methods to train robust supervised and unsupervised models. Adversarial noises can be even more capable of training robust classifiers than random noises by encouraging smooth predictions on both labeled and unlabeled examples that are adversarially affected. We will take a closer look at them in the context of semi-supervised methods in Section~\ref{sec:tsm}.

\subsubsection{GAN-based Representations}
In a GAN model, data are generated from the noises fed into its generator, and thus these noises can be viewed as the natural representations of data produced by the generator.  Considering the proved results \cite{arora2017generalization,qi2017loss} that many GAN variants have the generalized ability of generating data with indistinguishable distribution from that of real examples, the GAN representations are also complete for all real data.

However, there exists a challenge that given a real sample, we have to invert the generator to obtain the noise representation corresponding to the sample. Thus, an encoder is required that can directly output the noise from which the corresponding sample can be generated and thus represented.

For this purpose, the idea of adversarially training a generator and its corresponding encoder has been independently developed in Bidirectional Generative Adversarial Networks (BiGAN) \cite{donahue2016adversarial} and Adversarially Learned Inference (ALI) \cite{dumoulin2016adversarially}, respectively. The idea is later integrated into a regularized loss-sensitive GAN model with proved distributional consistency and generalizability to generate real data \cite{edraki2018generalized}.

\vspace{2mm}
\noindent{\bf BiGAN and ALI: Adversarial Representation Learning}
\vspace{2mm}

Formally, these methods aim to learn triple elements from a GAN model: 1) a generator $G: \mathcal Z \rightarrow \mathcal X$ mapping from a distribution $p(\mathbf z)$ of input noises $\mathcal Z$ to a distribution $p_g(\mathbf x)$ of generated samples $\mathcal X$; 2) an encoder $E: \mathcal X \rightarrow \mathcal Z$ mapping a sample $\mathbf x\in\mathcal X$ back to a noise $\mathbf z\in \mathcal Z$ such that ideally $G(\mathbf z)$ equals to $\mathbf x$, i.e., $E$ is the inverse of $G$; 3) a discriminator $D:\mathcal X \times \mathcal Z \rightarrow [0,1]$ that assigns a probability to distinguish a real pair $(\mathbf x, E(\mathbf x))$ from a fake pair $(G(\mathbf z),\mathbf z)$.

Compared with the classic GAN, there are two major differences. First, the encoder is the extra element for representation learning. Second, a discriminator has a joint sample-noise pair rather than a single sample as input to distinguish real from fake pairs.

The above triple elements can be jointly trained with a minimax objective

\begin{equation}\label{eq:bigan}
\begin{aligned}
\min_{G,E}\max_D V(D,E,G)
\end{aligned}
\end{equation}
where
\[
\begin{aligned}
V(D,E,G)&\triangleq \mathbb E_{\mathbf x \sim p(\mathbf x)} [\log D(\mathbf x, E(\mathbf x))] \\
&+ \mathbb E_{\mathbf z \sim p(\mathbf z)} [\log (1-D( G(\mathbf z), \mathbf z))]
\end{aligned}
\]
and $p(\mathbf x)$ is the real data distribution. This minimax problem can be solved by the alternating gradient based methods like in training the classic GAN \cite{goodfellow2014generative}.

Donahue et al. \cite{donahue2016adversarial} have proved in an ideal case, the resultant encoder $E$ inverts the generator $G$ almost everywhere, i.e., $E=G^{-1}$ (See Theorem 2 in \cite{donahue2016adversarial}). Moreover, it has also been shown that the joint training of $E$ and $G$ in (\ref{eq:bigan}) is performed by minimizing the $\ell_0$ loss of auto-encoders (See Theorem 3 of \cite{donahue2016adversarial}), which makes $E$ a desired representation model for its input samples.

\vspace{2mm}
\noindent{\bf More Related Works}
\vspace{2mm}

Besides BiGAN and ALI, there exist other hybrid methods jointly training auto-encoders and GANs to perform adversarial representation learning and inference in an integrated framework \cite{larsen2015autoencoding,makhzani2015adversarial,srivastava2017veegan,ulyanov2018takes,huang2018introvae}.

For example, Larsen et al. \cite{larsen2015autoencoding} use the intermediate representation from the GAN's discriminator to measure the similarity between reconstructed and input images as the reconstruction error to train the VAE. Alternatively, adversarial autoencoders \cite{makhzani2015adversarial} have been proposed to train the VAE by matching the aggregated posterior $q(\mathbf z)=\int_\mathbf x q(\mathbf z|\mathbf x) p(\mathbf x) d\mathbf x$ of the noises from the data distribution $p(\mathbf x)$ with that of prior distribution $p(\mathbf z)$. The match between distributions is performed by training a discriminator to tell $q(\mathbf z)$ apart from $p(\mathbf z)$ and guide the encoder $q(\mathbf z|\mathbf x)$ to produce the aggregated posterior indistinguishable from the prior.
In this way, the training of VAE is regularized to ensure the decoder yields a generative model that maps the given prior to the desired data distribution.

The marriage between VAE and GAN has also been explored to relieve the mode collapse problem. For example, Srivastava et al. \cite{srivastava2017veegan} train an encoder (called reconstructor in that paper) to invert the generator, and reduce the mode collapse of generated samples by having the distribution of encoded data match with the input Gaussian noise. The assumption is if mode collapses occur, it is unlikely for the reconstructor to map all generated samples back to the distribution of original Gaussian noises, and this results in a strong learning signal to train both generator and reconstructor.

Huang et al. \cite{huang2018introvae} have taken a further step by introducing an IntroAVE model with the posterior $q(\mathbf z|\mathbf x)$ as the discriminator directly to distinguish between real and fake data. Specifically, the posterior of $\mathbf z$ conditioned on real samples $\mathbf x$ is encouraged to match the prior $p(\mathbf z)$, while that of  $\mathbf z$ on the generated samples is supposed to deviate from $p(\mathbf z)$. Then, the generator can be trained to generate samples by matching the posterior with the prior. It has been shown that IntroAVE is able to generate the data indistinguishable from real samples.

\subsubsection{Disentangled Representations}\label{sec:dr}

Disentangling representations \cite{chen2016infogan} has been proposed to facilitate downstream tasks by providing interpretable and salient attributes to depict data. Bengio et al.~\cite{bengio2013representation} propose that a small subset of the latent variables in a disentangled representation ought to change
as data change in response to real-world events and transformations.

For example, a set of meaningful attributes, such as facial expressions, poses, eye colors, hairstyles, genders and even identities, can be separately allocated to disentangle facial images, and they can be extremely useful for solving future recognition problems without having to be exposed to some supervised data. This suggests good representations that are generalizable to natural supervised tasks ought to be as disentangle as possible to provide a rich set of factorized attributes to depict data.

\vspace{2mm}
\noindent{\bf InfoGAN: Disentangling GAN-based Representation}
\vspace{2mm}

The effort on disentangling representations has led to the InfoGAN \cite{chen2016infogan} and its variants \cite{higgins2016beta,jeon2018ib} in literature to train generative models that can create data from disentangled representations. Specifically, the InfoGAN assumes there are two types of noise variables fed into its generator: 1) a vector of incompressible noises $\mathbf z$, which do not factorize into any semantic representations and could be used by the generator in an entangled fashion as in the conventional GAN; 2) a vector of latent codes $\mathbf c$, which represent salient disentangled information about the generated sample $\mathbf x$ and will not be lost during the generative process.

Thus, the assumption of the InfoGAN is to maximize the mutual information between latent codes $\mathbf c$ and the generated samples $G(\mathbf z, \mathbf c)$ by combining combination these two types of noises. It should prevent the generator from ignoring the dependency on the latent codes that contain the salient knowledge about the generated samples. The mutual information $I(\mathbf c, G(\mathbf z, \mathbf c))$ is maximized over the generator $G$ to train the InfoGAN along with the minimax objective of the conventional GAN. A tractable variational lower bound of $I(\mathbf c, G(\mathbf z, \mathbf c))$ is derived by a surrogate distribution $q(\mathbf c|\mathbf x)$ to approximate the true posterior $p(\mathbf c|\mathbf x)$:
$$
I(\mathbf c, G(\mathbf z, \mathbf c)) \geq \mathbb E_{\mathbf c\sim p(\mathbf c), \mathbf x \sim G(\mathbf z, \mathbf c)}[\log q(\mathbf c|\mathbf x)] + H(\mathbf c)
$$
where $p(\mathbf c)$ is the prior distribution on latent codes and $H(\mathbf c)$ is its entropy. More details on the InfoGAN can be found in \cite{chen2016infogan}.

\vspace{2mm}
\noindent{\bf $\beta$-VAE: Disentangling VAE Representation}
\vspace{2mm}

The idea of disentangling representations has also been extended to other unsupervised models as well. Among them is $\beta$-VAE \cite{higgins2016beta}, which aims to disentangle the inferred posterior $q(\mathbf z|\mathbf x)$ by imposing a constraint on matching it to an isotropic Gaussian $p(\mathbf z)=\mathcal N(\mathbf 0, \mathbf I)$. It creates a latent information bottleneck on the inferred posterior by limiting its capacity. Such a regularization not only encourages a more efficient representation of data, but also disentangles the representations into independent factors due to the isotropic prior.

The following objective is maximized to train the VAE model
$$
\mathcal L(q(\mathbf z|\mathbf x),p(\mathbf x|\mathbf z)) = \mathbb E_{q(\mathbf z|\mathbf x)} p(\mathbf x|\mathbf z) - \beta D_{KL}(q(\mathbf z|\mathbf x)||p(\mathbf z))
$$
where the positive Lagrangian multiplier $\beta$ comes from the constraint $D_{KL}(q(\mathbf z|\mathbf x)||p(\mathbf z))<\epsilon$.

It is not hard to find when $\beta = 1$, the formulation reduces to the conventional VAE model.  As $\beta$ increases, a stronger constraint on the latent information bottleneck is enforced to control the capacity and conditional independence of the representation $q(\mathbf z|\mathbf x)$. A higher $\beta$ would trade off between the reconstruction fidelity of the $\beta$-VAE model and the disentanglement degree of the learned representations.

\vspace{2mm}
\noindent{\bf Disentanglement Metric}
\vspace{2mm}

To measure the degree of disentanglement of the learned representations, a {\em disentanglement metric score} \cite{higgins2016beta} is designed by the assumption that disentangled representations could enable robust classification of data based on their representations even using a simple classifier. A number of images are generated by fixing one of generative factors in the representations while randomly sampling all the others. Then a low capable linear classifier is used to identify this factor and the resultant accuracy is reported as the disentanglement metric score. Obviously, if the independence and interpretability property of the disentangled representations hold, the fixed factor should have a small variance, and thus the classifier ought to have high accuracy in identifying it and gives the high disentanglement score.

However, it is argued that a linear classifier could still be sensitive to hyperparameters and optimizers, and its disentanglement metric would suffer from a failure mode if only $K-1$ out of $K$ factors were disentangled. To address it, an alternative metric is proposed \cite{higgins2016beta} to directly use the variance of each dimension in the resultant representation as the indicator of the fixed factor, and apply a majority-vote classifier to predict the chosen factor. This avoids tuning optimization hyperparameters, as well as circumvents the failure mode of the other metric.

\vspace{2mm}
\noindent{\bf More Disentangled Representations}
\vspace{2mm}

Disentangling representations has been sought in many other generative models besides InfoGAN and $\beta$-VAE. The FactorVAE \cite{kim2018disentangling} proposes to minimize the Total Correlation (TC) $D_{KL}(q(\mathbf z)||\bar q(\mathbf z))$ between the aggregated posterior $q(\mathbf z)$ and its factorized form $\bar q(\mathbf z) = \prod_j q(\mathbf z_j)$, which measures the dependence for multiple random factors. Following the density-ratio trick, a discriminator is trained to distinguish samples between two posteriors and output the probability of a sample $\mathbf z$ being from the true aggregated posterior $q(\mathbf z)$. Then the factorized VAE is trained by minimizing the VAE lower bound along with the obtained TC. Compared with $\beta$-VAE, the FactorVAE avoids unnecessarily penalizing the mutual information $I(\mathbf x, \mathbf z)$ term, and thus yields better reconstruction of data while still sufficiently disentangling the representations of generative factors.

In addition, disentangling representations has also been studied in the context of semi-supervised methods \cite{kulkarni2015deep,karaletsos2015bayesian}, which will be reviewed in the next section.

\subsubsection{More Generative Models}

\vspace{2mm}
{\noindent\bf Flow-based Generative Models}
\vspace{2mm}

The flow-based generative models \cite{dinh2014nice,dinh2016density,kingma2018glow} map a random noise $\mathbf z$ drawn from a simple distribution (e.g., multi-variate Gaussian) to a data sample $\mathbf x$ through a series of bijective functions
$$
\mathbf x \overset {{f_1}} \longleftrightarrow \mathbf h_1 = f_1(\mathbf x) \overset {{f_2}} \longleftrightarrow \mathbf h_2 = f_2(\mathbf h_1) \cdots \overset {{f_K}} \longleftrightarrow \mathbf z = f_K(\mathbf h_{K-1})
$$
This sequence of invertible functions is called a flow. It allows us to compute the log-likelihood of $\mathbf x$ tractably by the change of variables formula as
$$
\log p(\mathbf x) = \log p(\mathbf z) + \sum_{i=1}^K \log |\det(\frac{d \mathbf h_i}{d \mathbf h_{i-1}})|.
$$

Three different types of invertible flow functions -- Actnorm, invertible convolution and affine coupling layer -- have been adopted \cite{kingma2018glow} to construct an one-step flow in a deep Generative fLOW (GLOW) model. A squeezing operator also defines a multi-scale structure with different levels of data abstraction in the GLOW \cite{dinh2016density}. Each step in GLOW has a log-determinant that can be easily computed as it has a triangular Jacobian matrix, and thus the resultant data loglikelihood can be maximized efficiently to train the model.

\vspace{2mm}
{\noindent\bf Self-Attention and Transformer}
\vspace{2mm}

The Transformer \cite{vaswani2017attention} has been proposed as an alternative to the recurrent neural networks, and it has stacked self-attention layers, as well as point-wise fully connected layers and positional encoding to capture the dependency between input and output sequences in its encoder and decoder components.

The self-attention is the key. Each embedding in a sequence is mapped to a tuple of query, key and value. Then the output at each position is a sum of the values weighted by the similarity between the current query and the keys of the sequence. A multi-head attention is often adopted to linearly project the queries, keys and values multiple times with different projection weights, and the resultant outputs from these linear projections are concatenated and projected to the final result.

Besides the  self-attention, each layer in the encoder and decoder contains a fully
connected feed-forward network applied to each position separately and identically. Positional encoding by sine and cosine functions is also added to each embedding, which provides information about the positions in the sequence. Transformer has become a powerful unsupervised representation of word embedding in natural language tasks, and more details about the Transformer and its application can be found in \cite{vaswani2017attention,devlin2018bert}.

\subsection{Self-Supervised Methods}\label{sec:self}

A large variety of self-supervisory signals have been proposed to train unsupervised representations as well.
We will start by reviewing the autoregressive models as a large category of self-supervised models, and proceed to learn self-supervised representations for images and videos. We will focus on basic ideas and principles for self-supervised representation learning in this survey.  For a more complete review of self-supervised learning on multimodal audio-visual data \cite{owens2016ambient,arandjelovic2018objects,korbar2018cooperative}, readers can refer to \cite{jing2019self}.

\subsubsection{Autoregressive Models}

One of categories of self-supervised models are trained by predicting the context, missing or future data, and they are often referred to as auto-regressive models. 
Among them are PixelRNN \cite{oord2016pixel}, PixelCNN \cite{van2016conditional,salimans2017pixelcnn++}, and Transformer \cite{vaswani2017attention}.
They can generate useful unsupervised representations since the contexts from which the unseen parts of data are predicted often depend on the same shared latent representations.

\vspace{2mm}
{\noindent\bf PixelRNN}
\vspace{2mm}

Specifically, in the PixelRNN \cite{oord2016pixel}, an image is divided into a regular grid of small patches, and a recurrent architecture is built to predict the features of the current patch based on its context. Three variants of RNNPixels are proposed to generate the sequence of image patches in different ways: Row LSTM, Diagonal BiLSTM and Multi-Scale PixelRNN.

For the Row LSTM \cite{oord2016pixel}, an image is generated row by row from top to bottom, and the context of a patch is roughly a triangle above the patch. In contrast, the Diagonal LSTM scans an image diagonally from a corner
at the top and reaches the opposite corner at the bottom, and thus it has a diagonal context. The Multi-Scale PixelRNN \cite{oord2016pixel} is composed of an unconditional PixelRNN and one or more PixelRNN layers. An unconditional PixelRNN is first applied to generate a smaller image subsampled from the original one, and then a conditional PixelRNN layer takes the smaller image as input to generate the original larger image. Multiple layers of conditional PixelRNN layers can be stacked to progressively generate the original image from the low to high resolutions.

\vspace{2mm}
{\noindent\bf PixelCNN}
\vspace{2mm}

A disadvantage of the Row and Diagonal LSTM is the high computational cost as the feature of each patch must be computed sequentially. This can be avoided by using a convolutional structure to compute the features of all patches at once. Masked convolutions are used to avoid the violation of conditional dependence only on the previous rather than future context. Compared with the PixelRNN with a potentially unbounded range of dependency, the PixelCNN \cite{van2016conditional} comes at a cost of limiting the context of each patch to a bounded receptive field. Thus multiple convolutional layers can be stacked to increase the context size.

On the other hand, gated activations have been introduced to the PixelCNN \cite{van2016conditional}. This results in a Gated PixelCNN that is able to model more complex interdependency between different patches. Moreover, the Gated PixelCNN is augmented with a horizontal stack conditioned on the current row so far, as well as a vertical stack dependent on all previous rows. By combining the outputs of both stacks, the blind spot can be avoided in the receptive field.  


\vspace{2mm}
{\noindent\bf Contrastive Predictive Coding and Instance Discrimination}
\vspace{2mm}

Auto-regressive models can be used as a decoder in the auto-encoder architecture, where they are forced to output powerful representations useful for predicting the future patches. This enables us to train representations in an auto-regressive fashion without accessing any labeled data.

Contrastive Predictive Coding (CPC) \cite{oord2018representation} has made a notable effort on training such auto-regressive models. It aims to maximize the mutual information $I(\mathbf c, \mathbf x)$ between the latent representations of the context $\mathbf c$ and the future sample $\mathbf x$, and thus more accurate future predictions can be made by maximally sharing information through the sequence. More details about CPC and its application in training auto-regressive models and learning unsupervised features can be found in \cite{oord2018representation}.

The idea of CPC has also inspired the non-parametric instance discrimination \cite{wu2018unsupervised}. By sampling a subset of samples into a memory bank \cite{wu2018unsupervised}, queue \cite{he2019momentum} or merely minibatch \cite{chen2020simple}, it trains an unsupervised representation by distinguishing positive pairs of augmented samples from negative ones. Formally, it minimizes the following contrast loss

\begin{equation}\label{eq:id}
\mathcal L = -\sum_{u,u'}\log \dfrac{\exp(s(u,u'))}{\sum_{u,v}\exp(s(u,v))}
\end{equation}
where $u,u'$ denote a positive pair of instances augmented from the same sample, $u,v$ are a pair of instances augmented from two random samples, and $s$ denotes a similarity function. Thus, this loss defines a non-parametric softmax loss to discriminate positive pairs against negative ones through their similarities.

More recently, together with stronger \cite{wang2021contrastive} and multi-crop \cite{caron2020unsupervised} augmentations, the unsupervised representation learned by the contrastive learning has achieved almost the same top-1 accuracy on ImageNet as its fully supervised counterpart with ResNet-50 \cite{wang2021contrastive}.  This demonstrates a noteworthy milestone for the great potentials of unsupervised learning in absence of labeled data.

\subsubsection{Image Representations}

In addition to autoregressive models,  self-supervised methods explore the other forms of self-supervised signals to train deep neural networks. These self-supervised signals can be directly derived from data themselves without having to manually label them.

{\bf \noindent Contexts.} For example, Doersch et al.~\cite{doersch2015unsupervised} use the relative positions of two randomly sampled patches from an image as self-supervised information to train the model. Pathak et al. \cite{pathak2016context} train a context encoder to generate the contents of missing parts from their surroundings by minimizing a combination of pixel-wise reconstruction error and an adversarial loss. Mehdi and Favaro~\cite{noroozi2016unsupervised} propose to train a convolutional neural network by solving Jigsaw puzzles.

{\bf \noindent Colorization.} Image colorization has also been used in a self-supervised task to train convolutional networks in literature \cite{zhang2016colorful,larsson2016learning}. Zhang et al.~\cite{zhang2017split} present a cross-channel auto-encoder by reconstructing a subset of data channels from another subset with the cross-channel features being concatenated as data representation.

{\bf \noindent Surrogate classes, targets and clustering.} Dosovitskiy et al.~\cite{dosovitskiy2014discriminative} train CNNs by classifying a set of surrogate classes, each of which is formed by applying various transformations to an individual image. In contrast, Bojanowski et al.~\cite{bojanowski2017unsupervised} use Noise As Target (NAT) by jointly learning the representation and assigning each sample to one of a fixed set of target values. Instead, Caron~\cite{caron2018deep} et al. train a DeepCluster model by iteratively clustering features and using the resultant representations to update the network.

{\bf\noindent Counting, motion and rotations.} Noroozi et al.~\cite{noroozi2017representation} learn counting features that satisfy equivalence relations between downsampled and tiled images. Egomotion \cite{agrawal2015learning} has also been used as a self-supervisory signal to model the representation of visual elements present in consecutive images to find their correspondences when an agent moves in an environment. Gidaris et al.~\cite{gidaris2018unsupervised} train neural networks by classifying image rotations in a discrete set. It learns a special case of transformation-equivariance representations as the learned representation ought to encode the information about them by equivarying with the applied rotations.


\subsubsection{Video Representations}

The idea of self-supervision has also been employed to train feature representations for videos by exploring the underlying temporal information. For example, the Arrow of Time (AoT) \cite{wei2018learning} has been used as the supervisory signal to learn the representations of videos for both high-level semantics and low-level physics, while avoiding artificial cues from the video production rather than the physical world.

The order of a sequence of frames can also supervise the training of video representations to capture the spatiotemporal information \cite{misra2016shuffle}. To this end, the Tuple verification approach is proposed to train a CNN model by extracting the representation of individual frames and determining whether a randomly sampled tuple of frames is in the correct order to disambiguate directional confusion in video clips.

A disentangled representation of images has also been proposed by leveraging the temporal coherence between video frames. A DrNet model \cite{denton2017unsupervised} is trained to factorize each frame into a stationary content representation and a time-varying pose representation with an adversarial loss. It assumes that the pose representation should carry no information about video identity, and the adversarial loss prevents the pose features from being discriminative from one video to another. The DrNet can learn powerful content and pose representations that can be combined to generate frames further into future than existing approaches \cite{denton2017unsupervised}.


\section{Semi-Supervised Methods}\label{sec:ssm}

In this section, we will review the semi-supervised methods \cite{wang2009semi,tang2007typicality,tang2008integrated,song2006video} from two different perspectives.

\begin{itemize}
\item {\bf Semi-supervised generative models.} In Section~\ref{sec:ssgm} The semi-supervised auto-encoders, GANs and disentangled representations will be reviewed in echoing their unsupervised counterparts. We will show how these semi-supervised generative models could be derived from the corresponding unsupervised generative models, shedding us some light on the intrinsic connection between the unsupervised and semi-supervised methods.
\item {\bf Teacher-Student models.} This is a large category of semi-supervised models that have achieved the state-of-the-art performances in literature, where a single or an ensemble of teacher models are trained to predict on unlabeled examples and the predicted labels are used to supervise the training of a student model. We will review various genres of teacher models -- noisy teachers, teacher ensemble and adversarial teachers -- in Section~\ref{sec:tsm}, and show how they could be trained against various noise and/or adversarial mechanisms to build more robust semi-supervised models.
\end{itemize}
We also provide an evaluation of different semi-supervised methods in Section~\ref{sec:ssl_eval} in the appendix.

\subsection{Semi-Supervised Generative Models}\label{sec:ssgm}

In this section, we will review a large variety of semi-supervised generative models.

\subsubsection{Semi-Supervised Auto-Encoders}

Kingma et al. \cite{kingma2014semi} extend the unsupervised variational auto-encoders to two forms of semi-supervised models.

The first latent-feature discriminative model (M1) is quite straightforward. On top of the latent representation $\mathbf z$ of a sample $\mathbf x$ by a VAE model, a classifier is trained to predict its label. While the VAE is trained on both the labeled and the unlabeled part of a training set, the classifier is trained based on labeled examples.

The second generative semi-supervised model (M2) is more complex. In addition to the latent representation $\mathbf z$, a sample $\mathbf x$ is generated by another class variable $y$, which is latent for a unlabeled $\mathbf x$ or seen for a labeled one. The data is explained by a generative process considering the additional class variable:
$$
p(y) = {\rm Cat}(y|\boldsymbol\pi), p(\mathbf z) = \mathcal N(\mathbf z|\mathbf 0, \mathbf I), p_\theta (\mathbf x|y, \mathbf z) = f_\theta(\mathbf x, y, \mathbf z)
$$
where $p(y)$ is a multinomial distribution for the class prior.

Unlike the VAE, the M2 introduces a pair of variational posteriors to infer $\mathbf z$ and $y$:
$$
q_\phi(\mathbf z|y,\mathbf x) = \mathcal N(\mathbf z|\boldsymbol \mu_\phi(y,\mathbf x),\boldsymbol \sigma_\phi^2(\mathbf x)), q_\phi(y|\mathbf x) = {\rm Cat}(y|\boldsymbol \pi_\phi(\mathbf x)).
$$
Then the joint posterior over $\mathbf z$ and $y$ can be inferred by $q_\phi(\mathbf z,y)=q_\phi(\mathbf z|y,\mathbf x)q_\phi(y|\mathbf x)$.

Among them, $q_\phi(y|\mathbf x)$ can be used as the classifier to predict the label of a test sample. To train the M2, two cases are considered \cite{kingma2014semi} to derive the variational lower bound of the marginal distribution  $p_\theta(\mathbf x, y)$ for labeled pairs $(\mathbf x, y)$ and $p_\theta(\mathbf x)$ for unlabeled samples $\mathbf x$, respectively. Combining the two bounds results in a maximum loglikelihood problem.

However, an additional classification cost ought to be added to the final objective function so that the classifier $q_\phi(y|\mathbf x)$ is trained with both labeled and unlabeled examples. Similar to the VAE, reparameterization trick is used to perform the back-propagation \cite{kingma2013auto}.

Finally, M1 and M2 can be combined by learning the M2 using the embedded representation $\mathbf z_1$ from a M1 model. The M2 model has its own latent representation $\mathbf z_2$ along with a label variable $y$ for each sample. This results in a two-layer deep generative model to generate $\mathbf z_1$ from $(\mathbf z_2,y)$ and $\mathbf x$ from $\mathbf z_1$ successively: $p_\theta(\mathbf x, y, \mathbf z_1, \mathbf z_2) = p(y)p(\mathbf z_2)p_\theta(\mathbf z_1|y,\mathbf z_2)p_\theta(\mathbf x|\mathbf z_1)$.

In addition to the M1 and M2 models and the hybrid, the efforts on introducing supervision information into the variational auto-encoders have been made in literature \cite{maaloe2016auxiliary,sonderby2016ladder,narayanaswamy2017learning} in different ways. Later on, we will review how to disentangle representations from the semi-supervised VAEs by partially specifying graphical dependency between a subset of random variables \cite{narayanaswamy2017learning} in order to factorize and interpret data variations.

\subsubsection{Semi-Supervised GANs}

The GANs have also been adopted to enable the semi-supervised learning from two different perspectives.
One of them considers to train a $K+1$ classifier with $K$ given labels to classify and a fake class to represent generated samples. It explores the distribution of unlabeled examples by treating them as belonging to the first $K$ real classes, and a  {\em feature matching} trick is used to unleash competitive performances \cite{salimans2016improved}.

On the contrary, the other paradigm views the generator of a learned GAN model as the (local) parameterization of the data manifold, so that the label invariance can be characterized over the manifold along its tangents. This is closely related with the Laplace-Beltrami operator that is merely approximated by the graph Laplacian in classic graph-based semi-supervised models.

We will review these two paradigms of semi-supervised GANs below.

\vspace{2mm}
{\noindent \bf Training $K+1$ Classifiers with Feature Matching}
\vspace{2mm}

Salimans et al. \cite{salimans2016improved} propose the improved techniques to train the semi-supervised GANs. By putting real and generated samples together, it trains a classifier to label each sample to one of $K$ real classes or a fake class. All unlabeled data are classified to real examples for one of the first $K$ classes, while the generated examples are classified to fake examples.  The conventional classification cost is defined over labeled data, which is combined with the unsupervised GAN loss to train the model.

Moreover, a trick called {\em feature matching} has to be used to train the generator. Instead of training the generator by maximizing the likelihood of its generated examples being classified to the $K$ real classes, it is trained by minimizing the discrepancy between the features of the real and the generated samples extracted from an intermediate layer of the classifier. This trick has played a critical role in delivering the competitive performances in training the semi-supervised GANs \cite{salimans2016improved}.

\vspace{2mm}
{\noindent\bf Pursuit of Label Invariance via Local GANs }
\vspace{2mm}

The graph Laplacian has been widely used to characterize the change of the labels over the samples connected in a graph.  Minimizing the graph Laplacian can make smooth predictions over the labels between the connected similar samples. While the graph is used to approximate the unknown data manifold, the graph Laplacian is indeed an approximate to the Laplace-Beltrami operator over the underlying data manifold.

In \cite{qi2018global}, a notable effort has been made to learn {\em localized} GAN that defines a local generator $G(\mathbf x, \mathbf z)$ around each sample $\mathbf x$ with $\mathbf z$. This gives rise to the local coordinates around each sample $\mathbf x$ over the data manifold in which $\mathbf x$ is the origin, i.e., $G(\mathbf x, \mathbf 0) = \mathbf x$. In this way, the entire data manifold can be covered by a family of local coordinates. It allows us to define the gradient of a classification function $f(\mathbf x)$ over the manifold as
\[
\begin{aligned}
\nabla^G_\mathbf x f\triangleq \nabla_\mathbf z f(G(\mathbf x,\mathbf z)) |_{\mathbf z=\mathbf 0}
= \mathbf J_\mathbf x^T \nabla_\mathbf x f(\mathbf x)
\end{aligned}
\]
where $\mathbf J_\mathbf x$ is the Jacobian matrix of $G(\mathbf x,\mathbf z)$ at $\mathbf z=\mathbf 0$.

With these notations, it can be revealed that the functional gradient over the manifold is closely connected with the Laplace-Beltrami operator $\triangle f \triangleq {\rm div} (\nabla^G_\mathbf x f)$ such that
\[
\int_\mathcal M \|\nabla^G_\mathbf x f\|^2 d P_\mathcal X
= \int_\mathcal M f \triangle f d P_\mathcal X.
\]
Therefore, one can directly calculate the Laplace-Beltrami operator without the approximate graph-based Laplacian that is often used in classic semi-supervised methods \cite{zhu2005semi}.

Then the semi-supervised classifier $p(\mathbf y|\mathbf x)$ is trained by  encouraging the label invariance over the data manifold by minimizing
$$
\sum_{k=1}^K \mathbb E_{p(\mathbf x)} \|\nabla^G_\mathbf x \log p(y=k|\mathbf x)\|^2
$$
along with the loss of the semi-supervised  GANs \cite{salimans2016improved}.

Moreover, the localized GAN allows us to explain the mode collapse of the generator from a geometric point of view as the manifold being local collapsed into a lower dimensionality. Then
an orthogonal constraint on the Jacobian matrix can be imposed to train the generator and prevent it from collapsing on the manifold.

\subsubsection{Semi-Supervised Disentangled Representations}

\vspace{2mm}
{\noindent\bf Inverse Graphics Networks}
\vspace{2mm}

The Deep Convolutional Inverse Graphics Network (DC-IGN) \cite{kulkarni2015deep} implements a semi-supervised variational auto-encoder model by engineering a vision model as inverse graphics. In other words, it aims to learn a collection of `` graphics codes" by which images can be transformed and rendered like in a graphics program. These graphics codes are viewed as disentangled representations of images.

DC-IGN is built on top of a VAE model, but is trained in a semi-supervised fashion. The learned representations are disentangled into few extrinsic variables such as azimuth angle, elevation angle and azimuth of light sources, along with a number of intrinsic variables depicting identity, shape, expression and surface textures. In a mini-batch, only one of factors is varied with all other others are fixed, generating the images with only one active transformation corresponding to the chosen factor that are fed forward through the network. The other variables corresponding to inactive transformations are clamped to their mean. The gradients of error signals are backpropagated through the network, while the gradients corresponding to the inactive transformations are forced to their difference from the mean over the mini-batch, and this could train the encoder such that all the information about the active transformation would be concentrated on the chosen variable.

The DC-IGN is semi-supervised to engineer inverse graphics as training images with various transformations are available from 3D face and chair datasets.
We also note that a number of inverse graphics models \cite{kulkarni2014inverse,jampani2015informed,mansinghka2013approximate} have been proposed to train disentangled representations. Among them are deep lambertian networks \cite{tang2012deep} that assume a Lambertian reflectance model and implicitly construct the 3D representation, and transforming auto-encoders \cite{hinton2011transforming,tieleman2014optimizing} that use a domain-specific decoder to reconstruct images, as well as \cite{loper2014opendr} with an approximate differentiable renderer to explicitly capture the relationship between changes in model parameters and image observations.

\vspace{2mm}
{\noindent \bf Disentangling Semi-Supervised VAEs}
\vspace{2mm}

In \cite{narayanaswamy2017learning}, a generalized form of semi-supervised VAEs is proposed to disentangle interpretable variables from the latent representations. It compiles the graphical model for modeling a general dependency on observed and unobserved latent variables with neural networks, and a stochastic computation graph \cite{schulman2015gradient} is used to infer with and train the resultant generative model.

For this purpose, importance sampling estimates are used to maximize the lower bound of both the supervised and semi-supervised likelihoods. By expanding each stochastic node into a subgraph, the stochastic computation graph is built to train the resultant model. Specifically, a distribution type and a neural network of parameter function are specified for each node in both the generative and inference models. The reparameterization trick is adopted to sample the unsupervised and semi-supervised variables, and the weight of the importance sampling is calculated from the joint probability of all semi-supervised variables.

This model enables us to flexibly specify the dependencies on the disentangled representations to interpret data variations, and leave the rest unspecified ones to be learned in an entangled fashion.

\subsection{Teacher-Student Models}\label{sec:tsm}
The idea behind teacher-student models for semi-supervised learning is to obtain a single or an ensemble of teachers, and use the predictions on unlabeled examples as targets to supervise the training of a student model. Consistency between the teacher and the student is maximized to improve the student's performance and stability on classifying unlabeled samples.

Various ways of training the teacher and maximizing the consistency between the teacher and the student lead to a variety of the semi-supervised models of this category.

Specifically, applying random noises to the input and hidden layers of models can be traced back to \cite{bishop1995training,reed1992regularization,srivastava2014dropout}, which have been shown to be equivalent to adding extra regularization terms to the objective function. In a teacher-student method, a {\em noisy teacher} is obtained by feeding noisy samples into a corrupted model, and the prediction bias is minimized to train the model between the teacher and the student ($\Gamma$-model \cite{rasmus2015semi}), or between two corrupted copies of the model ($\Pi$-model \cite{laine2016temporal}).

The idea is extended to convene an {\em ensemble of teachers} temporally over epochs to guide the training of their student. The exponential moving average of their predictions is used to improve the accuracy of predicted labels by the teacher ensemble on unlabeled examples (Temporal Ensembling \cite{laine2016temporal}). Alternatively, the exponential weighted average can be made over model parameters to form the predictions made by the teacher ensemble ({\em Mean Teacher} \cite{tarvainen2017mean}). Both methods rely on random noises added to input samples and model parameters respectively to improve the robustness of exploring unlabeled data when imposing the consistency between the teacher and student models.

Rather than adding random noises, adversarial examples are calculated that would maximally change the predicted labels by a student model. This yields an {\em adversarial teacher}, and the student is trained and updated by minimizing the deviation from the adversarial examples by the teacher. This yields the virtual Adversarial Training (VAT), which has achieved the state-of-the-art performance on semi-supervised learning.

In the following, we will elaborate on different teacher-student methods.

\subsubsection{Noisy Teachers: $\Gamma$ and $\Pi$ Models}
Both $\Gamma$ and $\Pi$ models are developed on the belief that a robust model ought to have stable predictions  under any random transformation of data and perturbations to the model \cite{sajjadi2016regularization}. This could push the decision boundary apart from training examples, and make the model insensitive to the noises on the data and the model parameters. Thus, random noises and perturbations are added into the inputs and the parameters of a student model to form a noisy teacher, and the deviation from the predictions by the teacher is minimized to train the student model.

Specifically, the $\Gamma$-model \cite{rasmus2015semi} has a multi-layered latent representation $\mathbf z^{(l)}$ of each layer $l$, and uses an auto-encoder to obtain an estimated $\hat{\mathbf z}^{(l)}$ by denoising from the corrupted $\tilde {\mathbf z}^{(l)}$. Then the sum of squared errors between the (batch-normalized) estimate and the clean latent representations over layers
$$
\sum_{l=1}^L \lambda_l \|\hat{\mathbf z}^{(l)}-{\mathbf z}^{(l)}\|^2
$$
is minimized to train the clear student model weighted by positive hyperparameter coefficients $\lambda_l$ across different layers.

On the contrary, $\Pi$-model \cite{laine2016temporal} is simplified by minimizing the difference between noisy outputs. In the context of semi-supervised learning problem, given a labeled or unlabeled sample $\mathbf x$ , it is corrupted by some noise and fed into the model perturbed by random dropout and pooling schemes \cite{sajjadi2016regularization}. This process is run twice, yielding two versions of its outputs $\mathbf y'$ and $\mathbf y''$. Then, the squared error between them is minimized to encourage the consistency between noisy outputs, combined with the classification cost on labeled examples to train the model. Unlike $\Gamma$-model that matches a clean and a corrupted representation, $\Pi$-model runs the corrupted branch twice to match noisy outputs.

However, both models rely on random noises to explore their resilience against noisy inputs and perturbed models, which would be ineffective in finding a competent teacher to train the robust models. Thus, an ensemble of teachers are tracked over epochs to form a more capable teacher model, resulting in the following {\em temporal ensembling} \cite{laine2016temporal} and {\em mean teacher} \cite{tarvainen2017mean} methods.


\subsubsection{Teacher Ensemble: Temporal Ensembling and Mean Teacher}

Temporal Ensembling \cite{laine2016temporal} and Mean Teacher \cite{tarvainen2017mean} are similar to each other in tracking an ensemble of models over time to have a better teacher model.
However, they differ in maintaining an exponential moving average over the predictions (temporal ensembling) by or the parameters (mean teacher) of the tracked models.

Formally, consider a model $\mathbf y = f_\theta(\mathbf x, \eta)$ parameterized by $\theta$ that outputs the prediction $\mathbf y$ for an input $\mathbf x$ under some noises $\eta$ added to the model parameters and/or the input.

For the temporal ensembling, at each epoch, the target prediction on a given sample $\mathbf x$ is updated in an Exponential Moving Average (EMA) fashion online as
$$
\mathbf y' \leftarrow \alpha \mathbf y' + (1-\alpha) f_\theta(\mathbf x, \eta)
$$
with a positive smoothing coefficient $\alpha$. The resultant EMA prediction is further normalized to construct a target $\mathbf y$ for training the model by minimizing
$$
\mathbb E_{\mathbf x, \eta}\|f_\theta (\mathbf x, \eta) - \mathbf y\|^2
$$
Again, this objective is combined with the classification cost over mini-batches to train the model $\theta$ corrupted with noise $\eta$. Since it is expensive to update the predictions over individual examples for every iteration, their target values are  updated only once per epoch, making the information from earlier models being incorporated into training the model  at a slower pace.

Contrary to temporal ensembling, the mean teacher keeps an EMA over the model parameters rather than individual predictions
$$
\theta' \leftarrow \alpha \theta' + (1-\alpha) \theta
$$
with the parameters $\theta$ of the current student model. Then the student model is updated by minimizing over $\theta$
$$
\mathbb E_{\mathbf x, \eta,\eta'}\|f_\theta (\mathbf x, \eta) - f_{\theta'} (\mathbf x, \eta')\|.
$$

While both temporal ensembling and mean teacher track a collection of previous models to predict the teacher's targets to supervise the training process, they still rely on adding {\em random} noises to train stable models with consistent predictions. It has been revealed that a locally isotropic output distribution around a sample cannot be achieved by training the model against randomly drawn noises without knowing the model's vulnerability to adversarial noises \cite{kurakin2016adversarial}.
This motivates an alternative method by using adversarial teachers \cite{miyato2018virtual} to supervise the training process.

\subsubsection{MixtureMatch and FixMatch Teachers}
MixMatch and FixMatch teachers represent another category of student-teacher model for semi-supervised learning.

In MixMatch \cite{berthelot2019mixmatch}, the current model makes a prediction on each unlabeled sample, which is linearly combined with the groundtruth label of anther example. This results in a predicted label on a mixed example on the segment of the unlabeled and labeled samples. This mixed example will be added to augment the training set to update the current model.

FixMatch \cite{sohn2020fixmatch} further simplifies MixMatch, and achieves even better performance.  It applies a stronger and a weaker augmentation to a unlabeled sample, and predicts the labels on two augmentations. It then trains by fixing the label on the weaker augmentation, and using it to teach on the stronger augmentation. In other words,
it seek to minimize the deviation between the labels on two augmentations, but only backpropagate its errors through the stronger one.  It is based on the assumption that the predicted label on the weaker augmentation is more likely to be true than that on the stronger augmentation, and it represents a {\em hierarchical augmentation} strategy for semi-supervised learning.

\subsubsection{Adversarial Teachers: Virtual Adversarial Training}

Adversarial training has been used to regularize a model and make it robust against adversarial examples \cite{kurakin2016adversarial,szegedy2013intriguing}.  Specifically, the model is trained to make a smooth prediction along an adversarial direction of input examples. This approach has been extended to Virtual Adversarial Training (VAT) \cite{miyato2018virtual}, where an adversarial direction can be sought around {\em unlabeled} data, along which the model is the most greatly altered. This allows to train the model in a semi-supervised fashion.

Formally, consider a labeled or an unlabeled example $\mathbf x$, and a parameterized model with a conditional distribution $p_\theta (\mathbf y|\mathbf x)$ of the output label. The VAT finds the most adversarial direction $\mathbf r_{adv}(\mathbf x)$ on $\mathbf x$ by
$$
\mathbf r_{adv}(\mathbf x) = \arg\max_{\|\mathbf r\|_2\leq \epsilon} D[p_\theta(\mathbf y|\mathbf x),p_\theta(\mathbf y|\mathbf x+\mathbf r)]
$$
with a divergence measure $D$ between two distributions, where the adversarial direction is sought within a radius $\epsilon$ around the sample.

Then an adversarial loss is minimized train the model
$$
\min_{\theta}\mathbb E_{\mathbf x}  D[p_\theta(\mathbf y|\mathbf x),p_\theta(\mathbf y|\mathbf x+\mathbf r_{adv}(\mathbf x))]
$$
over both labeled and unlabeled examples, together with the minimization of the classification cost.

The adversarial direction $\mathbf r_{adv} (\mathbf x)$ can be found in a closed form as the first dominant eigenvector of the Hessian matrix of $D[p_\theta(\mathbf y|\mathbf x),p_\theta(\mathbf y|\mathbf x+\mathbf r)]$ as a function of $\mathbf r$ at $\mathbf r=0$, which in turn allows a fast power iteration algorithm to solve $\mathbf r_{adv} (\mathbf x)$. This can be easily integrated into the stochastic gradient method to iteratively update $\theta$ over mini-batches.

\section{Domain Adaptation}\label{sec:domain}
We will review the domain adaptation problem in both unsupervised and semi-supervised fashion.

\subsection{Unsupervised Domain Adaptation}

One of interesting applications of the GANs is to adapt the learned representations and models from a source to a target domain.  Specifically, for the unsupervised domain adaptation,  a set of labeled source examples $S=\{(\mathbf x_i,\mathbf y_i)|i=1,\cdots,n\}$ are sampled from the distribution $p_S$ of a source domain, while there are another set of unlabeled examples $T=\{\mathbf x_i|i=1,\cdots,m\}$ from the distribution $p_T$ of a target domain.
Then the goal of the unsupervised domain adaptation is to learn a classifier $f$ that has a low risk $R_T={\rm Pr}_{(\mathbf x,y)\sim p_T}(f(\mathbf x)\neq y)$ on the target distribution. We categorize the unsupervised domain adaptation into the unsupervised method, since the target domain contains no supervision information although the source domain is supervised.

There are several different approaches to the unsupervised domain adaptation problem. Here we focus on reviewing the adversarial learning methods that are well related with the GAN models by leveraging the property that it can generate samples with an indistinguishable distribution from the target samples.

As outlined in \cite{tzeng2017adversarial}, there are three design choices in developing an unsupervised domain adaptation algorithm: 1) tying weights: whether the weights are shared across the representation models for the source and target domains; 2) base model: whether a discriminative or generative model is adapted from the source to target domain; 3) adversarial objectives that are used to train the models.

Different choices have resulted in various models.

\vspace{2mm}
{\noindent\bf Adversarial Discriminative Domain Adaptation}
\vspace{2mm}

Adversarial Discriminative Domain Adaptation (ADDA) \cite{tzeng2017adversarial} unties the weights of the representation models for source and target domain. Instead, it learns two separate models $M_S$ and $M_T$ to map source and target samples to their respective representations. First, a classifier $f$ is trained on top of the representation $M_S$ based on the labeled examples from the source domain:
$$
\min_{M_S,f} \mathbb E_{(\mathbf x, y)\sim p_S} \ell(f(M_S(\mathbf x)),y)
$$
where $\ell$ is the classification error on a labeled example.

Then $M_S$ is fixed, and the target representation model $M_T$ is trained so that both models output consistent distributions that match each other. A GAN-based objective is used to achieve this by learning a domain discriminator $D$ that distinguishes a source representation from its target counterpart,
$$
\max_D \mathbb E_{\mathbf x \sim p_S} \log D(M_S(\mathbf x)) + \mathbb E_{\mathbf x \sim p_T} (1-\log D(M_T(\mathbf x))).
$$

An adversarial loss is then minimized to train the target representation $M_T$ by confusing the domain discriminator that the representations generated by $M_T$ comes from the source domain:
$$
\max_{M_T} \mathbb E_{\mathbf x \sim p_T} \log D(M_T(\mathbf x)).
$$

The discriminator $D$ and the target representation $M_T$ are optimized iteratively to convergence. Then, a test sample $\mathbf x$ is classified by $f(M_T(\mathbf x))$ based on the trained classifier $f$ and the target model $M_T$.

\vspace{2mm}
{\noindent\bf Gradient Reversal Layer}
\vspace{2mm}

Unlike the ADDA, the Gradient Reversal Layer (GRL) model \cite{ganin2016domain} chooses to tie the weights for the source and target representations (i.e., $M_S=M_T=M$). The classifier $f$, the shared representation $M$, and the domain discriminator $D$ will be trained jointly.

It introduces the following regularizer over the shared $M$ and the domain discriminator $D$
\[
\begin{aligned}
\max_D\min_M R(D,M) &\triangleq \mathbb E_{\mathbf x \sim p_S} \log D(M(\mathbf x)) \\
&+ \mathbb E_{\mathbf x \sim p_T} (1-\log D(M(\mathbf x))).
\end{aligned}
\]
In other words, a shared representation $M$ is learned to map samples, no matter from the source or the target domain, to the same distribution such that $D$ cannot distinguish them.

This regularizer is combined with the classification loss, yielding the joint optimization problem
\begin{equation}\label{eq:GRL}
\max_D \min_{M,f} \mathbb E_{(\mathbf x, y)\sim p_S} \ell(f(M(\mathbf x)),y) + R(D,M).
\end{equation}

Compared with ADDA, the classifier is jointly trained with the representation, and it optimizes the true minimax objective that is vulnerable to the vanishing gradient \cite{tzeng2017adversarial}.

\subsection{Semi-Supervised Domain Adaptation}

The boundary between the unsupervised and the semi-supervised domain adaptations becomes blurred when additional labeled examples are available on the target domain.
For example, in the GRL, the classification loss (\ref{eq:GRL}) can be minimized with not only the labeled source examples but also the labeled target examples.

Alternatively, Pixel-Level Domain Adaptation (PixelDA) \cite{bousmalis2017unsupervised} chooses to directly adapt source images $\mathbf x\sim p_S$ to their target counterparts with a GAN generator $G(\mathbf x, \mathbf z)$ for a sampled noise $\mathbf z$ to match with the target distribution $p_T$. Then a classifier can be trained by combining the labeled adapted images $\{(G(\mathbf x,\mathbf z),y)|\mathbf (\mathbf x,y)\sim p_S\}$ and the labeled target images $\{(\mathbf x,y)|(\mathbf x,y)\sim p_T\}$ in a semi-supervised fashion. Additional content similarity loss can also be minimized to utilize the prior knowledge regarding the image adaptation process.

Moreover, a two-stream architecture has been proposed \cite{rozantsev2018beyond} to train two networks for the source and target domains simultaneously. It does not attempt to directly enforce domain invariance since domain invariant features could undermine the discriminator power of the learned classifiers. Instead, it explicitly models the domain shift by modeling both the similarity and the difference between the source and the target data.

Specifically, it trains two network streams separately on the labeled data from two domains. A weight regularizer is introduced by minimizing the difference between the weights of two network streams up to a linear transformation. This encourages two related streams to model the domain invariance while admitting the presence of the difference between domains. Then, the domain discrepancy can be minimized over the representations of source and target samples. This could be implemented by minimizing the Maximum Mean Discrepancy (MMD) \cite{tzeng2014deep,long2015learning,gretton2007kernel,long2013transfer} in a kernel space. In the meantime, the idea of GRL \cite{ganin2016domain} can also be applied to train a domain classifier that ought to perform poorly when the representations for two domains become indistinguishable.

\subsection{More Related Works}

There exist other variants of unsupervised domain adaptation methods based on the adversarial or non-adversarial training. For example, Domain Confusion \cite{tzeng2015simultaneous} proposes an objective under which the two untied representations are trained to map onto a uniform distribution by viewing two domains identically. CoGAN \cite{liu2016coupled} trains two GANs that generate the source and target images respectively. The domain-invariance is achieved by tying high-level parameters of the two GANs and a classifier is trained based on the output of the discriminator.

\section{Emerging Topics and Future Directions}\label{sec:related}
Now we will discuss emerging topics on unsupervised and semi-supervised learning and their future directions.

%
%
%


\subsection{Transformation Equivariance vs. Invariance}
A more theoretical topic lies on revealing the intrinsic relation between transformation equivariance and invariance in learning representations.
On the one hand, the pursuit of Transformation Equivariant Representation (TER) has been spotlighted as one of critical criteria \cite{hinton2011transforming} that achieves the state-of-the-art performances in unsupervised learning \cite{zhang2019aet,qi2019avt}. However, it is also important and necessary to apply the transformation invariance to train discriminative networks with labeled data in supervised tasks for recognizing images and objects.

At first glance, it looks like a dilemma to enforce two criteria simultaneously, but they actually co-exist well in the celebrated convolutional neural networks underpinning the great success of deep learning -- the convolutional feature maps equivary to the translations while the output predictions should be invariant under various transformations \cite{krizhevsky2012imagenet}.  The recent efforts \cite{cohen2016group} on generalizing translation equivariance to generic transformations also present great potentials of training more powerful representations and discriminative models atop to address small data challenges \cite{zhang2019aet,qi2019avt,cohen2016group}.

However, a deep understanding of the relationship between {\em transformation equivariance} and {\em transformation invariance} is still lacking to bridge the gap between training unsupervised and supervised models. While there is no doubt on the fundamental roles of transformation equivariance and invariance in unsupervised and supervised model, we still do not know the best way to integrate them in a coherent manner.

Indeed, the unsupervised representation learning concerns more on the generalizability to new tasks, while the supervised tasks are more interested in discriminative power for given tasks. How can the pursuits of transformation equivariance and invariance be suitably combined to reach better balance between generalization and discrimination?
Should we still separate the unsupervised learning of transformation equivariant representation from the supervised training for transformation invariant classifiers as in the CNNs? We believe insightful answers to these questions could lead to more transformative and efficient way to integrate both principles to address small data challenges for new tasks emerging everyday. This is a fundamental question we would like to answer in future.

\subsection{Unsupervised vs. Supervised Network Pretraining}

Pretraining of deep networks are often a critical step before they are finetuned on new tasks. For example, we often pretrain a deep network on ImageNet and fine-tune it on the other image datasets for various downstream tasks such as image classification, object detection and semantic segmentation.  While it has achieved huge successes, it is often limited to supervised pretraining on labeled datasets, and could result in inevitable gaps between the prelabeled datasets and the downstream problems.  For example, the prelabeled and the target datasets are often annotated with different labels. Even worse, they could focus on various tasks from image-level classification to localization of objects.  In such a setting, supervised pretraining may not be a natural solution to pretraining a deep network for unseen future tasks.

Fortunately, unsupervised training of deep networks can better generalize to downstream tasks without relying on pre-labeled datasets. It has great advantages in not only avoiding potential gaps to downstream tasks but also leveraging much larger unlabeled datasets.

%



%
\subsubsection{Unsupervised Pretraining for Future Tasks}

The results of unsupervised methods on cross-dataset tasks have shown impressive results. As shown in Table~\ref{tab:places}, the unsupervised models pretrained on the unlabeled ImageNet dataset have comparable  performances to the fully supervised models trained with the Places labels. Moreover, a unsupervised deep network pretrained on ImageNet achieves better performance on object detection task than its supervised pre-trained counterpart.
This demonstrates that the unsupervised pre-training is a promising alternative to the supervised pretraining approach.


This is not surprising. Indeed, the ability of unsupervised methods is more than that.  With more network capacities and larger datasets, we expect the unsupervised networks can deliver more impressive results. Unsupervised representations endowed with better generalizability to new tasks should benefit from a greater number of unlabeled data. To this end, more ambitious goals should be set to train more powerful unsupervised representations, and more challenging evaluation protocols on transfer learning scenarios should be considered in future.



%
%
\begin{table}
\caption{Error rates on CIFAR-10 when different numbers of labeled examples per class are used to train the supervised, the semi-supervised and the downstream classifiers for unsupervised representations.
For the unsupervised models, a convolutional block is trained with the labeled examples on top of the first twod blocks of NIN and 13-layer networks  pre-traine with all unlabeled data. We compare with both the fully supervised and the semi-supervised models.}\label{tab:ussl}
\centering
 \begin{tabular}{l|ccccc} \toprule
   &20&100&400&1000&5000\\ \midrule
Supervised conv &66.34&52.74 &25.81 &16.53 &6.93\\
Supervised non-linear &65.03&51.13 &27.17 &16.13 &7.92\\\midrule
Improved GAN \cite{salimans2016improved} & -- & -- & 18.63 & -- & -- \\
Temporal Ensembling \cite{laine2016temporal} &  -- & -- & 12.16 & -- & 5.60 \\
VAT+EntMin \cite{miyato2018virtual} &  -- & -- & 10.55 & -- & -- \\
$\Pi$ model & -- & 27.36 & 13.20 & -- & 6.06 \\
Localized GAN \cite{qi2018global} & -- & 17.44 & 14.23 & -- & -- \\
Mean Teacher \cite{tarvainen2017mean} & -- & 21.55 & 12.31 & -- & 5.94 \\\midrule
RotNet + conv (NIN) \cite{gidaris2018unsupervised}&35.37 &24.72&17.16&13.57 &8.05 \\
AET \cite{zhang2019aet} & 34.83&24.35 &16.28 &12.58 &7.82 \\
AVT  \cite{qi2019avt} &35.44&{24.26} &{15.97} &{12.57} &{7.75}\\
AVT (13-layers) \cite{qi2019avt}  & 26.20&18.44&13.56&10.86 &6.3\\\midrule
\end{tabular}
\end{table}
\subsubsection{Unsupervised Pretraining for Semi-Supervised Learning}

Another aspect of unsupervised training is its ability of exploring both labeled and unlabeled data in a semi-supervised fashion. First, a representation is trained on the unlabeled data, followed by training a classifier on top of the representation with a small number of labeled examples.
This differs from
the classic semi-supervised methods, in which a deep network is trained jointly on both labeled and unlabeled data.


In contrast, unsupervised training in the semi-supervised setting has its own advantage by
decoupling the unsupervised training of a base representation from the supervised training of a light-weighted classifier with few labeled examples.
This makes it more efficient to train unsupervised representations generalizable to new tasks, and could result in surprisingly competitive results.

Table~\ref{tab:ussl} compares the unsupervised methods with both fully supervised and semi-supervised models. Note that many compared semi-supervised and fully-supervised models are based on a 13-layer convolutional architecture \cite{laine2016temporal,tarvainen2017mean} on CIFAR-10, while the protocol adopted to compare the unsupervised models is often based on the NIN architecture. For the sake of a fair comparison, we also implement the same 13-layer architecture for the AVT model (the last row of Table~\ref{tab:ussl}), where the first two blocks of convolutions are unsupervised pretrained and the last block is trained with varying numbers of labeled data.
The result shows the promising potential of using unsupervised training in a semi-supervised way, since the AVT outperforms many existing semi-supervised models particularly when the number of labeled data is very small. We expect these unsupervised models could be further improved in future, and provide us with more flexibility and generalizability to handle new tasks with few labeled data in such a semi-supervised setting.

\subsection{Future Directions}

\begin{figure*}[t]
    \centering
    \begin{subfigure}[c]{0.32\textwidth}
        \includegraphics[width=\textwidth]{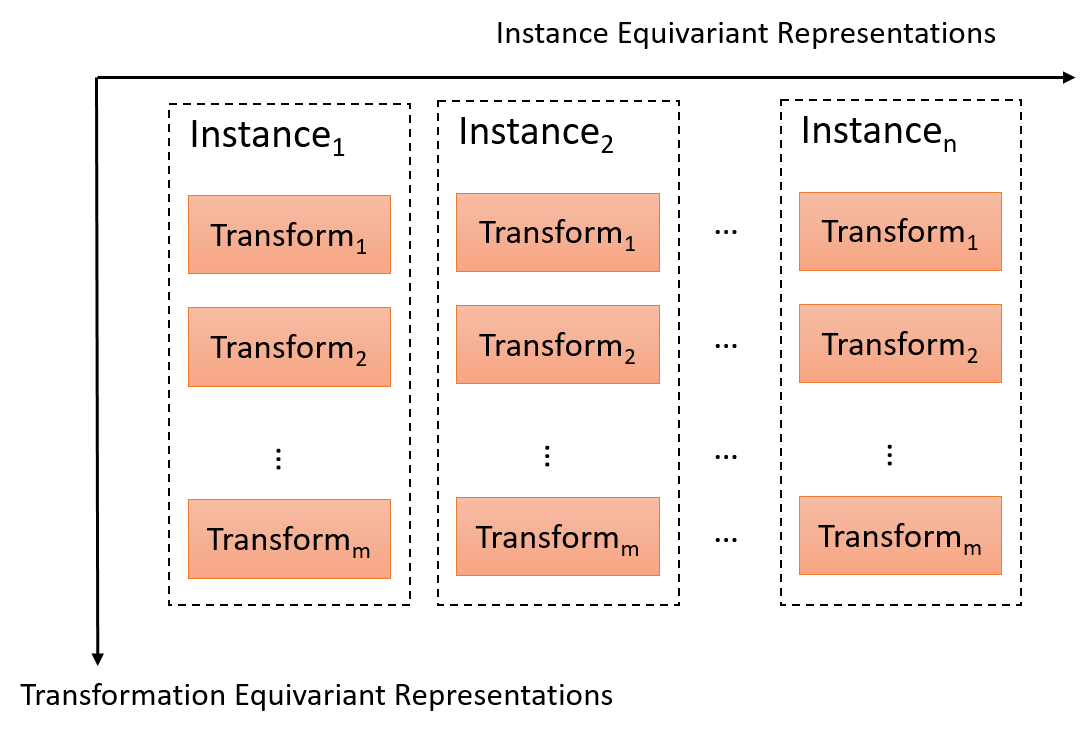}
        \caption{Unifying Transformation-and-Instance Equivariance}
    \end{subfigure}~
          \begin{subfigure}[c]{0.32\textwidth}
        \vspace{0mm}\includegraphics[width=\textwidth]{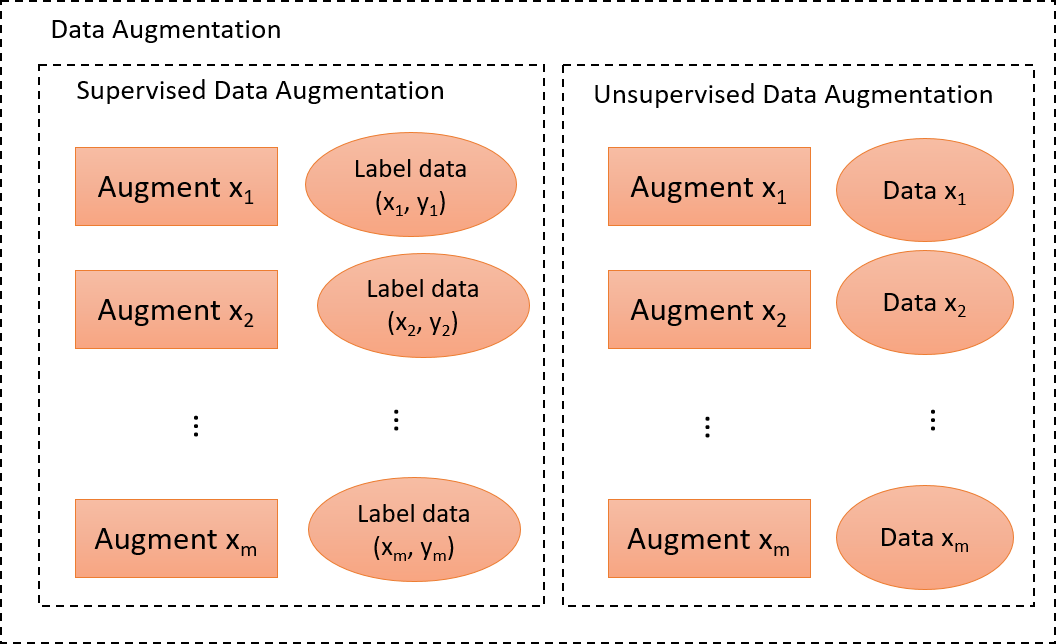}
        \vspace{0mm}\caption{Supervised vs. Unsupervised Augmentations}
    \end{subfigure}~
    \begin{subfigure}[c]{0.32\textwidth}
       \vspace{5mm} \includegraphics[width=\textwidth]{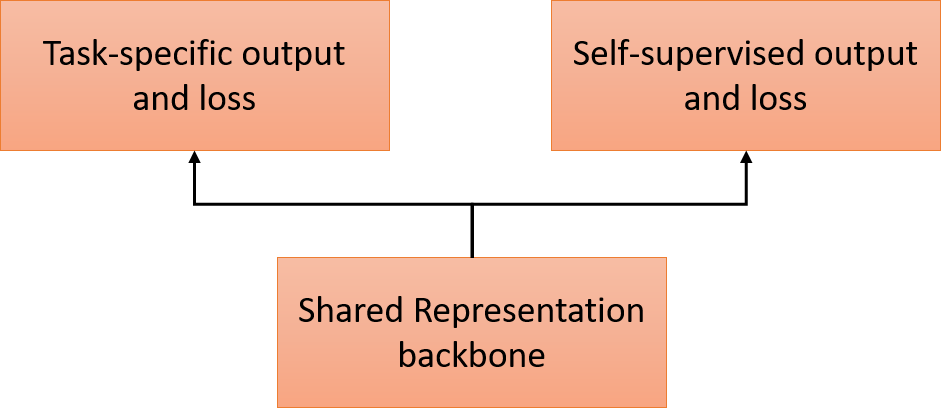}
        \vspace{5mm}\caption{Self-Supervised Learning As a Regularizer}
    \end{subfigure}
    \caption{Future directions for (a) unifying transformation and instance equivariant representation learning, (b) supervised vs. unsupervised data augmentations, and (c) self-supervised learning as a regularizer.}\label{fig:future}
\end{figure*}

Unsupervised and semi-supervised training of deep networks are closely related with many shared aspects of methodologies. As we have reviewed, unsupervised methods, such as Auto-Encoders, GANs and disentangled representations, all have their semi-supervised counterparts.  This is not surprising as they provide representation models that can be trained in both unsupervised and semi-supervised fashion.  This inspires us to explore both unsupervised and semi-supervised methods from an integrated point of view in the following directions as illustrated in Figure~\ref{fig:future}.

\subsubsection{Unifying Instance and Transformation Equivariances}

Instance discrimination \cite{wu2018unsupervised} and transformation prediction \cite{zhang2019aet,qi2019avt} have emerged as two large categories of unsupervised methods with leading performances in literature.  While the instance discrimination was originally inspired by the CPC reviewed in Section~\ref{sec:self}, the transformation prediction has been largely shaped by auto-encoding transformations on 2D images \cite{zhang2019aet} and 3D cloud points \cite{gao2019graphter}.

These two categories of methods approach the unsupervised learning from two distinctive dimensions of discrimination: {\em instance} \cite{wu2018unsupervised} and {\em transformation} \cite{zhang2019aet,qi2019avt}. Instance discrimination attempts to learn feature representations which ought to equivary to individual instances and thus are discriminative to distinguish one instance from another \cite{wu2018unsupervised}.  In contrast, transformation prediction seeks to learn the features that equivary to various transformations \cite{qi2019avt}, which can be a complex composition of spatial and non-spatial transformations \cite{wang2019enaet}. The learned representations should  be sufficiently discriminative from which applied transformations can be predicted.

This inspires us to learn feature representations that jointly {\em equivary} to both instances and their transformations.  Existing works we reviewed in this paper shed a light on unifying both directions from an information theoretic point of view.  For example, the contrastive loss in \cite{oord2018representation} was derived by maximizing the mutual information between the representations and their contexts (e.g., if they come from the same instance). On the other hand, the transformation equivariant representations \cite{qi2019avt} are also derived by maximizing the mutual information with the transformations as in Eq. (\ref{eq:avt}).  This leads to a natural choice to unify both equivariances by jointly maximizing the mutual information with both {\em instances} and {\em transformations}. While both equivariances have shown promising results in training expressive unsupervised representations, we believe an integrated solution has a greater potential that deserves our special attentions in future to close the performance gap to the fully supervised models as well as improve generalizability to unseen tasks.

\subsubsection{Semi-Supervised and Unsupervised Augmentations}

Data augmentation has become a standard preprocessing step in training deep networks since the introduction of Alexnet \cite{krizhevsky2012imagenet}.  It aims to augment the training set with more variations through different transformations such as spatial augmentations (e.g., random crops, mirror flips, random translations), color jittering, and random noises.

Augmentations can not only be applied to labeled examples to train supervised models (i.e., {\em supervised data augmentation}), but also play critical roles to explore variations in unsupervised setting (i.e., {\em semi-supervised data augmentation}).  In the review of semi-supervised methods, we have shown the roles of these augmentations in training robust semi-supervised models with noise-corrupted samples (cf. Section~\ref{sec:tsm}). A robust semi-supervised classifiers is trained to make consistent predictions over the augmented examples in ambient spaces \cite{sajjadi2016regularization} or along the tangent directions of the data manifold \cite{qi2018global}. The semi-supervised augmentations can also be added in an adversarial rather than random fashion to rectify the vulnerability of a model \cite{miyato2018virtual}.  A new trend has emerged to mix up the augmentations on labeled and unlabeled data in such a semi-supervised setting \cite{berthelot2019mixmatch}, or adopt a hierarchical data augmentation by fixing weakly augmented data to train strongly augmented ones \cite{sohn2020fixmatch}.


We also have unsupervised data augmentations that play a crucial role in training unsupervised models. 
For example, in instance discrimination \cite{wu2018unsupervised} and its predecessor Examplar-CNN \cite{dosovitskiy2014discriminative}, an unlabeled example is augmented into multiple versions under different transformations, and the unsupervised model is trained by distinguishing examples from one another up to these augmentations.  In contrast, the transformation prediction approaches such as the AET \cite{zhang2019aet} train unsupervised representations by directly predicting an applied transformation from a pair of examples augmented from the same instance.
Thus, these methods form two types of augmentations in unsupervised learning -- inter-instance augmentation for instance prediction, and intra-instance augmentation for transformation prediction. As mentioned in the last subsection, these two types of augmentations can be unified to jointly learn representations equivariant to both instances and transformations.

In future, we will also explore the potential of applying semi-supervised and unsupervised augmentations to various learning problems. For example, as discussed below, we can explore unsupervised augmentation in semi-supervised learning tasks as a self-trained regularizer.

Moreover, data augmentations can be combined with network augmentations to make the learned models more robust. For example, one can also add random or adversarial noises to networks on their weights and architectures to form an ensemble of augmented networks as in many teacher-student models we reviewed in Section~\ref{sec:tsm}.  The augmented networks can not only expose the potential vulnerability of network architectures (e.g., through adversarial dropout \cite{zhang2020wcp}), but also train more robust networks by exploring the change of network predictions in presence of noises on the weights.
We believe a combination of data and network augmentations can further improve the model robustness to learn more generalizable representations.


\subsubsection{Self-Supervision As a Regularizer}

Self-supervised learning we reviewed in Section~\ref{sec:self} not only constitutes a large category of unsupervised methods, but also shows impressive performances in various tasks as an unsupervised regularizer.  As illustrated in Figure~\ref{fig:future}(c), the idea is to train a shared backbone network jointly with a combination of target-dependent loss and a self-supervised regularizer.  Since the self-supervised loss only takes data as inputs, it does not rely on any task-specific labels and thus can be minimized regardless of the underlying task.

Here, we mention several learning tasks in which it is worth exploring the role of self-supervised regularization.
\vspace{2mm}

{\noindent \bf Self-Supervised Semi-Supervised Learning} It is natural to leverage self-supervised learning for semi-supervised tasks. Unsupervised and semi-supervised learning, which are the two main themes of this survey, converge here. A self-supervised loss can be applied to learn the instance \cite{wu2018unsupervised} and transformation \cite{qi2019avt} equivariances over unlabeled data, which can be combined with a supervised loss such as  cross-entropy on labeled data.  This seeks to {\em regularize} the training of classifiers by exploring such inter- and intra-instance variations in the learned representations under a composite of spatial, color, and temporal transformations and augmentations.  Exciting performances have been shown to train self-supervised semi-supervised models \cite{zhai2019s4l,wang2019enaet}. The current methods only use the self-supervised loss to regularize the training of the classifiers in an indirect fashion through the shared representations.  In future, more efforts are needed to understand the role of self-supervised regularizer on the predicted labels of the jointly trained classifiers, and we expect a theory can be developed to help us understand the self-supervised regularization in the supervised training, which can further improve performances.

\vspace{2mm}

{\noindent \bf Self-Supervised Domain Adaptation} Self-supervised regularizer has also been applied to bridge the domain gap \cite{carlucci2019domain,sun2019unsupervised} in addition to the methods reviewed in Section~\ref{sec:domain}. This combines the self-supervised loss such as jigsaw loss \cite{noroozi2016unsupervised} and transformation prediction loss \cite{gidaris2018unsupervised} to understand intrinsic data variations when adapting classifiers across different domains. It is based on the assumption that the unsupervised nature of a self-supervised task enables the learning of a common representation space across domains by aligning source and target samples regardless of their labels.  This will be able to generalize a source classifier to the target domain.  However, although it has demonstrated competitive performances \cite{carlucci2019domain,sun2019unsupervised}, we still need to devote more efforts to understand how the domain gap is closed by a common self-supervised task, as well as the relations with the other domain adaptation methods.

\vspace{2mm}

{\noindent \bf Self-Supervised Generative Adversarial Networks} Self-supervised regularization is also applied to train the GANs.  The idea of using rotation prediction task as a regularizer was presented in \cite{chen2019self} to self-train the GAN discriminators. It aims to combine the advantages of self-supervision and adversarial training to bridge the gap between conditional and unconditional GANs.  A more sophisticated Transformation GAN \cite{wang2020transformation} was proposed recently to explore the joint distribution of samples and their transformations. It extends the idea of the self-supervised AET \cite{zhang2019aet} by forcing the discriminator to distinguish between real and fake samples as well as their transformed copies.  This makes the self-trained GANs better generalizable in producing unseen samples of variations corresponding to different transformations.

\vspace{2mm}


The task-agnostic nature of self-supervised learning enables its wide applicability for many problems beyond the aforementioned tasks. In future, we can explore the application of self-supervision in more learning problems,
as well as seek a unified theory of self-supervised regularization for a large variety of learning tasks.



\section{Conclusions}\label{sec:concl}
In this paper, we review two large categories of small data methods -- unsupervised and semi-supervised methods. In particular, a large variety of generative models are reviewed, including auto-encoders, GANs, Flow-based models, and autoregressive models, in both supervised and semi-supervised categories. We also compare several emerging criteria and principles in training these models, such as transformation equivariance and invariance in training unsupervised and supervised representations, and the disentanglement of unsupervised and semi-supervised representations for factorized and interpretable deep networks. Unsupervised and semi-supervised domain adaptations have also been reviewed to reveal the recent progress on bridging the gaps between distributions of different domains in presence of unlabeled and labeled data, respectively. We also discuss the future directions to reveal the connections between unsupervised and semi-supervised learning.


%

%
%

\ifCLASSOPTIONcaptionsoff
  \newpage
\fi



{\footnotesize
\bibliographystyle{IEEEtran}
\bibliography{egbib,aet_egbib,small_data}

\begin{thebibliography}{100}
\providecommand{\url}[1]{#1}
\csname url@samestyle\endcsname
\providecommand{\newblock}{\relax}
\providecommand{\bibinfo}[2]{#2}
\providecommand{\BIBentrySTDinterwordspacing}{\spaceskip=0pt\relax}
\providecommand{\BIBentryALTinterwordstretchfactor}{4}
\providecommand{\BIBentryALTinterwordspacing}{\spaceskip=\fontdimen2\font plus
\BIBentryALTinterwordstretchfactor\fontdimen3\font minus
  \fontdimen4\font\relax}
\providecommand{\BIBforeignlanguage}[2]{{%
\expandafter\ifx\csname l@#1\endcsname\relax
\typeout{** WARNING: IEEEtran.bst: No hyphenation pattern has been}%
\typeout{** loaded for the language `#1'. Using the pattern for}%
\typeout{** the default language instead.}%
\else
\language=\csname l@#1\endcsname
\fi
#2}}
\providecommand{\BIBdecl}{\relax}
\BIBdecl

\bibitem{krizhevsky2012imagenet}
A.~Krizhevsky, I.~Sutskever, and G.~E. Hinton, ``Imagenet classification with
  deep convolutional neural networks,'' in \emph{Advances in neural information
  processing systems}, 2012, pp. 1097--1105.

\bibitem{he2016deep}
K.~He, X.~Zhang, S.~Ren, and J.~Sun, ``Deep residual learning for image
  recognition,'' in \emph{Proceedings of the IEEE conference on computer vision
  and pattern recognition}, 2016, pp. 770--778.

\bibitem{finn2017model}
C.~Finn, P.~Abbeel, and S.~Levine, ``Model-agnostic meta-learning for fast
  adaptation of deep networks,'' in \emph{Proceedings of the 34th International
  Conference on Machine Learning-Volume 70}.\hskip 1em plus 0.5em minus
  0.4em\relax JMLR. org, 2017, pp. 1126--1135.

\bibitem{jamal2018task}
M.~A. Jamal, G.-J. Qi, and M.~Shah, ``Task-agnostic meta-learning for few-shot
  learning,'' \emph{arXiv preprint arXiv:1805.07722}, 2018.

\bibitem{li2019learning}
X.~Li, Q.~Sun, Y.~Liu, Q.~Zhou, S.~Zheng, T.-S. Chua, and B.~Schiele,
  ``Learning to self-train for semi-supervised few-shot classification,'' in
  \emph{Advances in Neural Information Processing Systems}, 2019, pp.
  10\,276--10\,286.

\bibitem{hinton2011transforming}
G.~E. Hinton, A.~Krizhevsky, and S.~D. Wang, ``Transforming auto-encoders,'' in
  \emph{International Conference on Artificial Neural Networks}.\hskip 1em plus
  0.5em minus 0.4em\relax Springer, 2011, pp. 44--51.

\bibitem{zhang2019aet}
L.~Zhang, G.-J. Qi, L.~Wang, and J.~Luo, ``Aet vs. aed: Unsupervised
  representation learning by auto-encoding transformations rather than data,''
  \emph{arXiv preprint arXiv:1901.04596}, 2019.

\bibitem{fu2017recent}
Y.~Fu, T.~Xiang, Y.-G. Jiang, X.~Xue, L.~Sigal, and S.~Gong, ``Recent advances
  in zero-shot recognition,'' \emph{arXiv preprint arXiv:1710.04837}, 2017.

\bibitem{qi2018global}
G.-J. Qi, L.~Zhang, H.~Hu, M.~Edraki, J.~Wang, and X.-S. Hua, ``Global versus
  localized generative adversarial nets,'' in \emph{Proceedings of IEEE
  Conference on Computer Vision and Pattern Recognition (CVPR)}, 2018.

\bibitem{laine2016temporal}
S.~Laine and T.~Aila, ``Temporal ensembling for semi-supervised learning,''
  \emph{arXiv preprint arXiv:1610.02242}, 2016.

\bibitem{miyato2018virtual}
T.~Miyato, S.-i. Maeda, S.~Ishii, and M.~Koyama, ``Virtual adversarial
  training: a regularization method for supervised and semi-supervised
  learning,'' \emph{IEEE transactions on pattern analysis and machine
  intelligence}, 2018.

\bibitem{ren2018meta}
M.~Ren, E.~Triantafillou, S.~Ravi, J.~Snell, K.~Swersky, J.~B. Tenenbaum,
  H.~Larochelle, and R.~S. Zemel, ``Meta-learning for semi-supervised few-shot
  classification,'' \emph{arXiv preprint arXiv:1803.00676}, 2018.

\bibitem{ma2019affinitynet}
T.~Ma and A.~Zhang, ``Affinitynet: semi-supervised few-shot learning for
  disease type prediction,'' in \emph{Proceedings of the AAAI Conference on
  Artificial Intelligence}, vol.~33, 2019, pp. 1069--1076.

\bibitem{chen2018a}
\BIBentryALTinterwordspacing
W.-Y. Chen, Y.-C. Liu, Z.~Kira, Y.-C.~F. Wang, and J.-B. Huang, ``A closer look
  at few-shot classification,'' in \emph{International Conference on Learning
  Representations}, 2019. [Online]. Available:
  \url{https://openreview.net/forum?id=HkxLXnAcFQ}
\BIBentrySTDinterwordspacing

\bibitem{wu2018unsupervised}
Z.~Wu, Y.~Xiong, S.~X. Yu, and D.~Lin, ``Unsupervised feature learning via
  non-parametric instance discrimination,'' in \emph{Proceedings of the IEEE
  Conference on Computer Vision and Pattern Recognition}, 2018, pp. 3733--3742.

\bibitem{qi2019avt}
G.-J. Qi and et~al., ``Avt: Unsupervised learning of transformation equivariant
  representations by autoencoding variational transformations,'' 2019.

\bibitem{rasmus2015semi}
A.~Rasmus, M.~Berglund, M.~Honkala, H.~Valpola, and T.~Raiko, ``Semi-supervised
  learning with ladder networks,'' in \emph{Advances in Neural Information
  Processing Systems}, 2015, pp. 3546--3554.

\bibitem{tarvainen2017mean}
A.~Tarvainen and H.~Valpola, ``Mean teachers are better role models:
  Weight-averaged consistency targets improve semi-supervised deep learning
  results,'' in \emph{Advances in neural information processing systems}, 2017,
  pp. 1195--1204.

\bibitem{berthelot2019mixmatch}
D.~Berthelot, N.~Carlini, I.~Goodfellow, N.~Papernot, A.~Oliver, and C.~A.
  Raffel, ``Mixmatch: A holistic approach to semi-supervised learning,'' in
  \emph{Advances in Neural Information Processing Systems}, 2019, pp.
  5050--5060.

\bibitem{sohn2020fixmatch}
K.~Sohn, D.~Berthelot, C.-L. Li, Z.~Zhang, N.~Carlini, E.~D. Cubuk, A.~Kurakin,
  H.~Zhang, and C.~Raffel, ``Fixmatch: Simplifying semi-supervised learning
  with consistency and confidence,'' \emph{arXiv preprint arXiv:2001.07685},
  2020.

\bibitem{kurakin2016adversarial}
A.~Kurakin, I.~Goodfellow, and S.~Bengio, ``Adversarial examples in the
  physical world,'' \emph{arXiv preprint arXiv:1607.02533}, 2016.

\bibitem{szegedy2013intriguing}
C.~Szegedy, W.~Zaremba, I.~Sutskever, J.~Bruna, D.~Erhan, I.~Goodfellow, and
  R.~Fergus, ``Intriguing properties of neural networks,'' \emph{arXiv preprint
  arXiv:1312.6199}, 2013.

\bibitem{cohen2016group}
T.~Cohen and M.~Welling, ``Group equivariant convolutional networks,'' in
  \emph{International conference on machine learning}, 2016, pp. 2990--2999.

\bibitem{cohen2018intertwiners}
T.~S. Cohen, M.~Geiger, and M.~Weiler, ``Intertwiners between induced
  representations (with applications to the theory of equivariant neural
  networks),'' \emph{arXiv preprint arXiv:1803.10743}, 2018.

\bibitem{sabour2017dynamic}
S.~Sabour, N.~Frosst, and G.~E. Hinton, ``Dynamic routing between capsules,''
  in \emph{Advances in Neural Information Processing Systems}, 2017, pp.
  3856--3866.

\bibitem{lenssen2018group}
J.~E. Lenssen, M.~Fey, and P.~Libuschewski, ``Group equivariant capsule
  networks,'' \emph{arXiv preprint arXiv:1806.05086}, 2018.

\bibitem{cohen2018spherical}
T.~S. Cohen, M.~Geiger, J.~K{\"o}hler, and M.~Welling, ``Spherical cnns,''
  \emph{arXiv preprint arXiv:1801.10130}, 2018.

\bibitem{cohen2016steerable}
T.~S. Cohen and M.~Welling, ``Steerable cnns,'' \emph{arXiv preprint
  arXiv:1612.08498}, 2016.

\bibitem{goodfellow2014generative}
I.~Goodfellow, J.~Pouget-Abadie, M.~Mirza, B.~Xu, D.~Warde-Farley, S.~Ozair,
  A.~Courville, and Y.~Bengio, ``Generative adversarial nets,'' in
  \emph{Advances in neural information processing systems}, 2014, pp.
  2672--2680.

\bibitem{donahue2016adversarial}
J.~Donahue, P.~Kr{\"a}henb{\"u}hl, and T.~Darrell, ``Adversarial feature
  learning,'' \emph{arXiv preprint arXiv:1605.09782}, 2016.

\bibitem{dumoulin2016adversarially}
V.~Dumoulin, I.~Belghazi, B.~Poole, O.~Mastropietro, A.~Lamb, M.~Arjovsky, and
  A.~Courville, ``Adversarially learned inference,'' \emph{arXiv preprint
  arXiv:1606.00704}, 2016.

\bibitem{srivastava2017veegan}
A.~Srivastava, L.~Valkov, C.~Russell, M.~U. Gutmann, and C.~Sutton, ``Veegan:
  Reducing mode collapse in gans using implicit variational learning,'' in
  \emph{Advances in Neural Information Processing Systems}, 2017, pp.
  3308--3318.

\bibitem{larsen2015autoencoding}
A.~B.~L. Larsen, S.~K. S{\o}nderby, H.~Larochelle, and O.~Winther,
  ``Autoencoding beyond pixels using a learned similarity metric,'' \emph{arXiv
  preprint arXiv:1512.09300}, 2015.

\bibitem{makhzani2015adversarial}
A.~Makhzani, J.~Shlens, N.~Jaitly, I.~Goodfellow, and B.~Frey, ``Adversarial
  autoencoders,'' \emph{arXiv preprint arXiv:1511.05644}, 2015.

\bibitem{rifai2011contractive}
S.~Rifai, P.~Vincent, X.~Muller, X.~Glorot, and Y.~Bengio, ``Contractive
  auto-encoders: Explicit invariance during feature extraction,'' in
  \emph{Proceedings of the 28th International Conference on International
  Conference on Machine Learning}.\hskip 1em plus 0.5em minus 0.4em\relax
  Omnipress, 2011, pp. 833--840.

\bibitem{kingma2013auto}
D.~P. Kingma and M.~Welling, ``Auto-encoding variational bayes,'' \emph{arXiv
  preprint arXiv:1312.6114}, 2013.

\bibitem{vincent2008extracting}
P.~Vincent, H.~Larochelle, Y.~Bengio, and P.-A. Manzagol, ``Extracting and
  composing robust features with denoising autoencoders,'' in \emph{Proceedings
  of the 25th international conference on Machine learning}.\hskip 1em plus
  0.5em minus 0.4em\relax ACM, 2008, pp. 1096--1103.

\bibitem{vincent2010stacked}
P.~Vincent, H.~Larochelle, I.~Lajoie, Y.~Bengio, and P.-A. Manzagol, ``Stacked
  denoising autoencoders: Learning useful representations in a deep network
  with a local denoising criterion,'' \emph{Journal of machine learning
  research}, vol.~11, no. Dec, pp. 3371--3408, 2010.

\bibitem{arora2017generalization}
S.~Arora, R.~Ge, Y.~Liang, T.~Ma, and Y.~Zhang, ``Generalization and
  equilibrium in generative adversarial nets (gans),'' \emph{arXiv preprint
  arXiv:1703.00573}, 2017.

\bibitem{qi2017loss}
G.-J. Qi, ``Loss-sensitive generative adversarial networks on lipschitz
  densities,'' \emph{arXiv preprint arXiv:1701.06264}, 2017.

\bibitem{edraki2018generalized}
M.~Edraki and G.-J. Qi, ``Generalized loss-sensitive adversarial learning with
  manifold margins,'' in \emph{Proceedings of European Conference on Computer
  Vision (ECCV 2018)}, 2018.

\bibitem{ulyanov2018takes}
D.~Ulyanov, A.~Vedaldi, and V.~Lempitsky, ``It takes (only) two: Adversarial
  generator-encoder networks,'' in \emph{Thirty-Second AAAI Conference on
  Artificial Intelligence}, 2018.

\bibitem{huang2018introvae}
H.~Huang, R.~He, Z.~Sun, T.~Tan \emph{et~al.}, ``Introvae: Introspective
  variational autoencoders for photographic image synthesis,'' in
  \emph{Advances in Neural Information Processing Systems}, 2018, pp. 52--63.

\bibitem{chen2016infogan}
X.~Chen, Y.~Duan, R.~Houthooft, J.~Schulman, I.~Sutskever, and P.~Abbeel,
  ``Infogan: Interpretable representation learning by information maximizing
  generative adversarial nets,'' in \emph{Advances in neural information
  processing systems}, 2016, pp. 2172--2180.

\bibitem{bengio2013representation}
Y.~Bengio, A.~Courville, and P.~Vincent, ``Representation learning: A review
  and new perspectives,'' \emph{IEEE transactions on pattern analysis and
  machine intelligence}, vol.~35, no.~8, pp. 1798--1828, 2013.

\bibitem{higgins2016beta}
I.~Higgins, L.~Matthey, A.~Pal, C.~Burgess, X.~Glorot, M.~Botvinick,
  S.~Mohamed, and A.~Lerchner, ``beta-vae: Learning basic visual concepts with
  a constrained variational framework,'' 2016.

\bibitem{jeon2018ib}
I.~Jeon, W.~Lee, and G.~Kim, ``Ib-gan: Disentangled representation learning
  with information bottleneck gan,'' 2018.

\bibitem{kim2018disentangling}
H.~Kim and A.~Mnih, ``Disentangling by factorising,'' \emph{arXiv preprint
  arXiv:1802.05983}, 2018.

\bibitem{kulkarni2015deep}
T.~D. Kulkarni, W.~F. Whitney, P.~Kohli, and J.~Tenenbaum, ``Deep convolutional
  inverse graphics network,'' in \emph{Advances in neural information
  processing systems}, 2015, pp. 2539--2547.

\bibitem{karaletsos2015bayesian}
T.~Karaletsos, S.~Belongie, and G.~R{\"a}tsch, ``Bayesian representation
  learning with oracle constraints,'' \emph{arXiv preprint arXiv:1506.05011},
  2015.

\bibitem{dinh2014nice}
L.~Dinh, D.~Krueger, and Y.~Bengio, ``Nice: Non-linear independent components
  estimation,'' \emph{arXiv preprint arXiv:1410.8516}, 2014.

\bibitem{dinh2016density}
L.~Dinh, J.~Sohl-Dickstein, and S.~Bengio, ``Density estimation using real
  nvp,'' \emph{arXiv preprint arXiv:1605.08803}, 2016.

\bibitem{kingma2018glow}
D.~P. Kingma and P.~Dhariwal, ``Glow: Generative flow with invertible 1x1
  convolutions,'' in \emph{Advances in Neural Information Processing Systems},
  2018, pp. 10\,236--10\,245.

\bibitem{vaswani2017attention}
A.~Vaswani, N.~Shazeer, N.~Parmar, J.~Uszkoreit, L.~Jones, A.~N. Gomez,
  {\L}.~Kaiser, and I.~Polosukhin, ``Attention is all you need,'' in
  \emph{Advances in Neural Information Processing Systems}, 2017, pp.
  5998--6008.

\bibitem{devlin2018bert}
J.~Devlin, M.-W. Chang, K.~Lee, and K.~Toutanova, ``Bert: Pre-training of deep
  bidirectional transformers for language understanding,'' \emph{arXiv preprint
  arXiv:1810.04805}, 2018.

\bibitem{owens2016ambient}
A.~Owens, J.~Wu, J.~H. McDermott, W.~T. Freeman, and A.~Torralba, ``Ambient
  sound provides supervision for visual learning,'' in \emph{European
  conference on computer vision}.\hskip 1em plus 0.5em minus 0.4em\relax
  Springer, 2016, pp. 801--816.

\bibitem{arandjelovic2018objects}
R.~Arandjelovic and A.~Zisserman, ``Objects that sound,'' in \emph{Proceedings
  of the European Conference on Computer Vision (ECCV)}, 2018, pp. 435--451.

\bibitem{korbar2018cooperative}
B.~Korbar, D.~Tran, and L.~Torresani, ``Cooperative learning of audio and video
  models from self-supervised synchronization,'' in \emph{Advances in Neural
  Information Processing Systems}, 2018, pp. 7763--7774.

\bibitem{jing2019self}
L.~Jing and Y.~Tian, ``Self-supervised visual feature learning with deep neural
  networks: A survey,'' \emph{arXiv preprint arXiv:1902.06162}, 2019.

\bibitem{oord2016pixel}
A.~v.~d. Oord, N.~Kalchbrenner, and K.~Kavukcuoglu, ``Pixel recurrent neural
  networks,'' \emph{arXiv preprint arXiv:1601.06759}, 2016.

\bibitem{van2016conditional}
A.~Van~den Oord, N.~Kalchbrenner, L.~Espeholt, O.~Vinyals, A.~Graves
  \emph{et~al.}, ``Conditional image generation with pixelcnn decoders,'' in
  \emph{Advances in Neural Information Processing Systems}, 2016, pp.
  4790--4798.

\bibitem{salimans2017pixelcnn++}
T.~Salimans, A.~Karpathy, X.~Chen, and D.~P. Kingma, ``Pixelcnn++: Improving
  the pixelcnn with discretized logistic mixture likelihood and other
  modifications,'' \emph{arXiv preprint arXiv:1701.05517}, 2017.

\bibitem{oord2018representation}
A.~v.~d. Oord, Y.~Li, and O.~Vinyals, ``Representation learning with
  contrastive predictive coding,'' \emph{arXiv preprint arXiv:1807.03748},
  2018.

\bibitem{he2019momentum}
K.~He, H.~Fan, Y.~Wu, S.~Xie, and R.~Girshick, ``Momentum contrast for
  unsupervised visual representation learning,'' \emph{arXiv preprint
  arXiv:1911.05722}, 2019.

\bibitem{chen2020simple}
T.~Chen, S.~Kornblith, M.~Norouzi, and G.~Hinton, ``A simple framework for
  contrastive learning of visual representations,'' \emph{arXiv preprint
  arXiv:2002.05709}, 2020.

\bibitem{wang2021contrastive}
X.~Wang and G.-J.~Q. Qi, ``Contrastive learning with stronger augmentations,''
  in \emph{Submitted to International Conference on Learning Representations},
  2021, under review.

\bibitem{caron2020unsupervised}
M.~Caron, I.~Misra, J.~Mairal, P.~Goyal, P.~Bojanowski, and A.~Joulin,
  ``Unsupervised learning of visual features by contrasting cluster
  assignments,'' \emph{arXiv preprint arXiv:2006.09882}, 2020.

\bibitem{doersch2015unsupervised}
C.~Doersch, A.~Gupta, and A.~A. Efros, ``Unsupervised visual representation
  learning by context prediction,'' in \emph{Proceedings of the IEEE
  International Conference on Computer Vision}, 2015, pp. 1422--1430.

\bibitem{pathak2016context}
D.~Pathak, P.~Krahenbuhl, J.~Donahue, T.~Darrell, and A.~A. Efros, ``Context
  encoders: Feature learning by inpainting,'' in \emph{Proceedings of the IEEE
  Conference on Computer Vision and Pattern Recognition}, 2016, pp. 2536--2544.

\bibitem{noroozi2016unsupervised}
M.~Noroozi and P.~Favaro, ``Unsupervised learning of visual representations by
  solving jigsaw puzzles,'' in \emph{European Conference on Computer
  Vision}.\hskip 1em plus 0.5em minus 0.4em\relax Springer, 2016, pp. 69--84.

\bibitem{zhang2016colorful}
R.~Zhang, P.~Isola, and A.~A. Efros, ``Colorful image colorization,'' in
  \emph{European Conference on Computer Vision}.\hskip 1em plus 0.5em minus
  0.4em\relax Springer, 2016, pp. 649--666.

\bibitem{larsson2016learning}
G.~Larsson, M.~Maire, and G.~Shakhnarovich, ``Learning representations for
  automatic colorization,'' in \emph{European Conference on Computer
  Vision}.\hskip 1em plus 0.5em minus 0.4em\relax Springer, 2016, pp. 577--593.

\bibitem{zhang2017split}
R.~Zhang, P.~Isola, and A.~A. Efros, ``Split-brain autoencoders: Unsupervised
  learning by cross-channel prediction.''

\bibitem{dosovitskiy2014discriminative}
A.~Dosovitskiy, J.~T. Springenberg, M.~Riedmiller, and T.~Brox,
  ``Discriminative unsupervised feature learning with convolutional neural
  networks,'' in \emph{Advances in Neural Information Processing Systems},
  2014, pp. 766--774.

\bibitem{bojanowski2017unsupervised}
P.~Bojanowski and A.~Joulin, ``Unsupervised learning by predicting noise,''
  \emph{arXiv preprint arXiv:1704.05310}, 2017.

\bibitem{caron2018deep}
M.~Caron, P.~Bojanowski, A.~Joulin, and M.~Douze, ``Deep clustering for
  unsupervised learning of visual features,'' \emph{arXiv preprint
  arXiv:1807.05520}, 2018.

\bibitem{noroozi2017representation}
M.~Noroozi, H.~Pirsiavash, and P.~Favaro, ``Representation learning by learning
  to count,'' in \emph{The IEEE International Conference on Computer Vision
  (ICCV)}, 2017.

\bibitem{agrawal2015learning}
P.~Agrawal, J.~Carreira, and J.~Malik, ``Learning to see by moving,'' in
  \emph{Proceedings of the IEEE International Conference on Computer Vision},
  2015, pp. 37--45.

\bibitem{gidaris2018unsupervised}
S.~Gidaris, P.~Singh, and N.~Komodakis, ``Unsupervised representation learning
  by predicting image rotations,'' \emph{arXiv preprint arXiv:1803.07728},
  2018.

\bibitem{wei2018learning}
D.~Wei, J.~J. Lim, A.~Zisserman, and W.~T. Freeman, ``Learning and using the
  arrow of time,'' in \emph{Proceedings of the IEEE Conference on Computer
  Vision and Pattern Recognition}, 2018, pp. 8052--8060.

\bibitem{misra2016shuffle}
I.~Misra, C.~L. Zitnick, and M.~Hebert, ``Shuffle and learn: unsupervised
  learning using temporal order verification,'' in \emph{European Conference on
  Computer Vision}.\hskip 1em plus 0.5em minus 0.4em\relax Springer, 2016, pp.
  527--544.

\bibitem{denton2017unsupervised}
E.~L. Denton \emph{et~al.}, ``Unsupervised learning of disentangled
  representations from video,'' in \emph{Advances in Neural Information
  Processing Systems}, 2017, pp. 4414--4423.

\bibitem{wang2009semi}
M.~Wang, X.-S. Hua, T.~Mei, R.~Hong, G.~Qi, Y.~Song, and L.-R. Dai,
  ``Semi-supervised kernel density estimation for video annotation,''
  \emph{Computer Vision and Image Understanding}, vol. 113, no.~3, pp.
  384--396, 2009.

\bibitem{tang2007typicality}
J.~Tang, X.-S. Hua, G.-J. Qi, and X.~Wu, ``Typicality ranking via
  semi-supervised multiple-instance learning,'' in \emph{Proceedings of the
  15th ACM international conference on Multimedia}, 2007, pp. 297--300.

\bibitem{tang2008integrated}
J.~Tang, H.~Li, G.-J. Qi, and T.-S. Chua, ``Integrated graph-based
  semi-supervised multiple/single instance learning framework for image
  annotation,'' in \emph{Proceedings of the 16th ACM international conference
  on Multimedia}, 2008, pp. 631--634.

\bibitem{song2006video}
Y.~Song, G.-J. Qi, X.-S. Hua, L.-R. Dai, and R.-H. Wang, ``Video annotation by
  active learning and semi-supervised ensembling,'' in \emph{2006 IEEE
  International Conference on Multimedia and Expo}.\hskip 1em plus 0.5em minus
  0.4em\relax IEEE, 2006, pp. 933--936.

\bibitem{kingma2014semi}
D.~P. Kingma, S.~Mohamed, D.~J. Rezende, and M.~Welling, ``Semi-supervised
  learning with deep generative models,'' in \emph{Advances in neural
  information processing systems}, 2014, pp. 3581--3589.

\bibitem{maaloe2016auxiliary}
L.~Maal{\o}e, C.~K. S{\o}nderby, S.~K. S{\o}nderby, and O.~Winther, ``Auxiliary
  deep generative models,'' \emph{arXiv preprint arXiv:1602.05473}, 2016.

\bibitem{sonderby2016ladder}
C.~K. S{\o}nderby, T.~Raiko, L.~Maal{\o}e, S.~K. S{\o}nderby, and O.~Winther,
  ``Ladder variational autoencoders,'' in \emph{Advances in neural information
  processing systems}, 2016, pp. 3738--3746.

\bibitem{narayanaswamy2017learning}
S.~Narayanaswamy, T.~B. Paige, J.-W. Van~de Meent, A.~Desmaison, N.~Goodman,
  P.~Kohli, F.~Wood, and P.~Torr, ``Learning disentangled representations with
  semi-supervised deep generative models,'' in \emph{Advances in Neural
  Information Processing Systems}, 2017, pp. 5925--5935.

\bibitem{salimans2016improved}
T.~Salimans, I.~Goodfellow, W.~Zaremba, V.~Cheung, A.~Radford, and X.~Chen,
  ``Improved techniques for training gans,'' in \emph{Advances in Neural
  Information Processing Systems}, 2016, pp. 2234--2242.

\bibitem{zhu2005semi}
X.~Zhu, ``Semi-supervised learning with graphs,'' Ph.D. dissertation, 2005.

\bibitem{kulkarni2014inverse}
T.~D. Kulkarni, V.~K. Mansinghka, P.~Kohli, and J.~B. Tenenbaum, ``Inverse
  graphics with probabilistic cad models,'' \emph{arXiv preprint
  arXiv:1407.1339}, 2014.

\bibitem{jampani2015informed}
V.~Jampani, S.~Nowozin, M.~Loper, and P.~V. Gehler, ``The informed sampler: A
  discriminative approach to bayesian inference in generative computer vision
  models,'' \emph{Computer Vision and Image Understanding}, vol. 136, pp.
  32--44, 2015.

\bibitem{mansinghka2013approximate}
V.~K. Mansinghka, T.~D. Kulkarni, Y.~N. Perov, and J.~Tenenbaum, ``Approximate
  bayesian image interpretation using generative probabilistic graphics
  programs,'' in \emph{Advances in Neural Information Processing Systems},
  2013, pp. 1520--1528.

\bibitem{tang2012deep}
Y.~Tang, R.~Salakhutdinov, and G.~Hinton, ``Deep lambertian networks,''
  \emph{arXiv preprint arXiv:1206.6445}, 2012.

\bibitem{tieleman2014optimizing}
T.~Tieleman, \emph{Optimizing neural networks that generate images}.\hskip 1em
  plus 0.5em minus 0.4em\relax University of Toronto (Canada), 2014.

\bibitem{loper2014opendr}
M.~M. Loper and M.~J. Black, ``Opendr: An approximate differentiable
  renderer,'' in \emph{European Conference on Computer Vision}.\hskip 1em plus
  0.5em minus 0.4em\relax Springer, 2014, pp. 154--169.

\bibitem{schulman2015gradient}
J.~Schulman, N.~Heess, T.~Weber, and P.~Abbeel, ``Gradient estimation using
  stochastic computation graphs,'' in \emph{Advances in Neural Information
  Processing Systems}, 2015, pp. 3528--3536.

\bibitem{bishop1995training}
C.~M. Bishop, ``Training with noise is equivalent to tikhonov regularization,''
  \emph{Neural computation}, vol.~7, no.~1, pp. 108--116, 1995.

\bibitem{reed1992regularization}
R.~Reed, S.~Oh, and R.~Marks, ``Regularization using jittered training data,''
  in \emph{Neural Networks, 1992. IJCNN., International Joint Conference on},
  vol.~3.\hskip 1em plus 0.5em minus 0.4em\relax IEEE, 1992, pp. 147--152.

\bibitem{srivastava2014dropout}
N.~Srivastava, G.~Hinton, A.~Krizhevsky, I.~Sutskever, and R.~Salakhutdinov,
  ``Dropout: a simple way to prevent neural networks from overfitting,''
  \emph{The Journal of Machine Learning Research}, vol.~15, no.~1, pp.
  1929--1958, 2014.

\bibitem{sajjadi2016regularization}
M.~Sajjadi, M.~Javanmardi, and T.~Tasdizen, ``Regularization with stochastic
  transformations and perturbations for deep semi-supervised learning,'' in
  \emph{Advances in Neural Information Processing Systems}, 2016, pp.
  1163--1171.

\bibitem{tzeng2017adversarial}
E.~Tzeng, J.~Hoffman, K.~Saenko, and T.~Darrell, ``Adversarial discriminative
  domain adaptation,'' in \emph{Computer Vision and Pattern Recognition
  (CVPR)}, vol.~1, no.~2, 2017, p.~4.

\bibitem{ganin2016domain}
Y.~Ganin, E.~Ustinova, H.~Ajakan, P.~Germain, H.~Larochelle, F.~Laviolette,
  M.~Marchand, and V.~Lempitsky, ``Domain-adversarial training of neural
  networks,'' \emph{The Journal of Machine Learning Research}, vol.~17, no.~1,
  pp. 2096--2030, 2016.

\bibitem{bousmalis2017unsupervised}
K.~Bousmalis, N.~Silberman, D.~Dohan, D.~Erhan, and D.~Krishnan, ``Unsupervised
  pixel-level domain adaptation with generative adversarial networks,'' in
  \emph{The IEEE Conference on Computer Vision and Pattern Recognition (CVPR)},
  vol.~1, no.~2, 2017, p.~7.

\bibitem{rozantsev2018beyond}
A.~Rozantsev, M.~Salzmann, and P.~Fua, ``Beyond sharing weights for deep domain
  adaptation,'' \emph{IEEE transactions on pattern analysis and machine
  intelligence}, 2018.

\bibitem{tzeng2014deep}
E.~Tzeng, J.~Hoffman, N.~Zhang, K.~Saenko, and T.~Darrell, ``Deep domain
  confusion: Maximizing for domain invariance,'' \emph{arXiv preprint
  arXiv:1412.3474}, 2014.

\bibitem{long2015learning}
M.~Long, Y.~Cao, J.~Wang, and M.~I. Jordan, ``Learning transferable features
  with deep adaptation networks,'' \emph{arXiv preprint arXiv:1502.02791},
  2015.

\bibitem{gretton2007kernel}
A.~Gretton, K.~Borgwardt, M.~Rasch, B.~Sch{\"o}lkopf, and A.~J. Smola, ``A
  kernel method for the two-sample-problem,'' in \emph{Advances in neural
  information processing systems}, 2007, pp. 513--520.

\bibitem{long2013transfer}
M.~Long, J.~Wang, G.~Ding, J.~Sun, and P.~S. Yu, ``Transfer feature learning
  with joint distribution adaptation,'' in \emph{Proceedings of the IEEE
  international conference on computer vision}, 2013, pp. 2200--2207.

\bibitem{tzeng2015simultaneous}
E.~Tzeng, J.~Hoffman, T.~Darrell, and K.~Saenko, ``Simultaneous deep transfer
  across domains and tasks,'' in \emph{Proceedings of the IEEE International
  Conference on Computer Vision}, 2015, pp. 4068--4076.

\bibitem{liu2016coupled}
M.-Y. Liu and O.~Tuzel, ``Coupled generative adversarial networks,'' in
  \emph{Advances in neural information processing systems}, 2016, pp. 469--477.

\bibitem{gao2019graphter}
X.~Gao, W.~Hu, and G.-J. Qi, ``Graphter: Unsupervised learning of graph
  transformation equivariant representations via auto-encoding node-wise
  transformations,'' 2020.

\bibitem{wang2019enaet}
X.~Wang, D.~Kihara, J.~Luo, and G.-J. Qi, ``Enaet: Self-trained ensemble
  autoencoding transformations for semi-supervised learning,'' \emph{arXiv
  preprint arXiv:1911.09265}, 2019.

\bibitem{zhang2020wcp}
L.~Zhang and G.-J. Qi, ``Wcp: Worst-case perturbations for semi-supervised deep
  learning,'' in \emph{Proceedings of the IEEE Conference on Computer Vision
  and Pattern Recognition}, 2020.

\bibitem{zhai2019s4l}
X.~Zhai, A.~Oliver, A.~Kolesnikov, and L.~Beyer, ``S4l: Self-supervised
  semi-supervised learning,'' in \emph{Proceedings of the IEEE international
  conference on computer vision}, 2019, pp. 1476--1485.

\bibitem{carlucci2019domain}
F.~M. Carlucci, A.~D'Innocente, S.~Bucci, B.~Caputo, and T.~Tommasi, ``Domain
  generalization by solving jigsaw puzzles,'' in \emph{Proceedings of the IEEE
  Conference on Computer Vision and Pattern Recognition}, 2019, pp. 2229--2238.

\bibitem{sun2019unsupervised}
Y.~Sun, E.~Tzeng, T.~Darrell, and A.~A. Efros, ``Unsupervised domain adaptation
  through self-supervision,'' \emph{arXiv preprint arXiv:1909.11825}, 2019.

\bibitem{chen2019self}
T.~Chen, X.~Zhai, M.~Ritter, M.~Lucic, and N.~Houlsby, ``Self-supervised gans
  via auxiliary rotation loss,'' in \emph{Proceedings of the IEEE Conference on
  Computer Vision and Pattern Recognition}, 2019, pp. 12\,154--12\,163.

\bibitem{wang2020transformation}
J.~Wang, W.~Zhou, G.-J. Qi, Z.~Fu, Q.~Tian, and H.~Li, ``Transformation gan for
  unsupervised image synthesis and representation learning,'' in
  \emph{Proceedings of the IEEE Conference on Computer Vision and Pattern
  Recognition}, 2020.

\bibitem{oyallon2015deep}
E.~Oyallon and S.~Mallat, ``Deep roto-translation scattering for object
  classification,'' in \emph{Proceedings of the IEEE Conference on Computer
  Vision and Pattern Recognition}, 2015, pp. 2865--2873.

\bibitem{radford2015unsupervised}
A.~Radford, L.~Metz, and S.~Chintala, ``Unsupervised representation learning
  with deep convolutional generative adversarial networks,'' \emph{arXiv
  preprint arXiv:1511.06434}, 2015.

\bibitem{oyallon2017scaling}
E.~Oyallon, E.~Belilovsky, and S.~Zagoruyko, ``Scaling the scattering
  transform: Deep hybrid networks,'' in \emph{International Conference on
  Computer Vision (ICCV)}, 2017.

\bibitem{wang2015unsupervised}
X.~Wang and A.~Gupta, ``Unsupervised learning of visual representations using
  videos,'' in \emph{Proceedings of the IEEE International Conference on
  Computer Vision}, 2015, pp. 2794--2802.

\bibitem{krahenbuhl2015data}
P.~Kr{\"a}henb{\"u}hl, C.~Doersch, J.~Donahue, and T.~Darrell, ``Data-dependent
  initializations of convolutional neural networks,'' \emph{arXiv preprint
  arXiv:1511.06856}, 2015.

\bibitem{zhou2014learning}
B.~Zhou, A.~Lapedriza, J.~Xiao, A.~Torralba, and A.~Oliva, ``Learning deep
  features for scene recognition using places database,'' in \emph{Advances in
  neural information processing systems}, 2014, pp. 487--495.

\bibitem{oliver2018realistic}
A.~Oliver, A.~Odena, C.~A. Raffel, E.~D. Cubuk, and I.~Goodfellow, ``Realistic
  evaluation of deep semi-supervised learning algorithms,'' in \emph{Advances
  in Neural Information Processing Systems}, 2018, pp. 3235--3246.

\bibitem{krizhevsky2009learning}
A.~Krizhevsky, ``Learning multiple layers of features from tiny images,''
  Citeseer, Tech. Rep., 2009.

\bibitem{netzer2011reading}
Y.~Netzer, T.~Wang, A.~Coates, A.~Bissacco, B.~Wu, and A.~Y. Ng, ``Reading
  digits in natural images with unsupervised feature learning,'' 2011.

\end{thebibliography}
}
%
%
%

%

\begin{IEEEbiography}[{\includegraphics[width=1in,height=1.25in,clip,keepaspectratio]{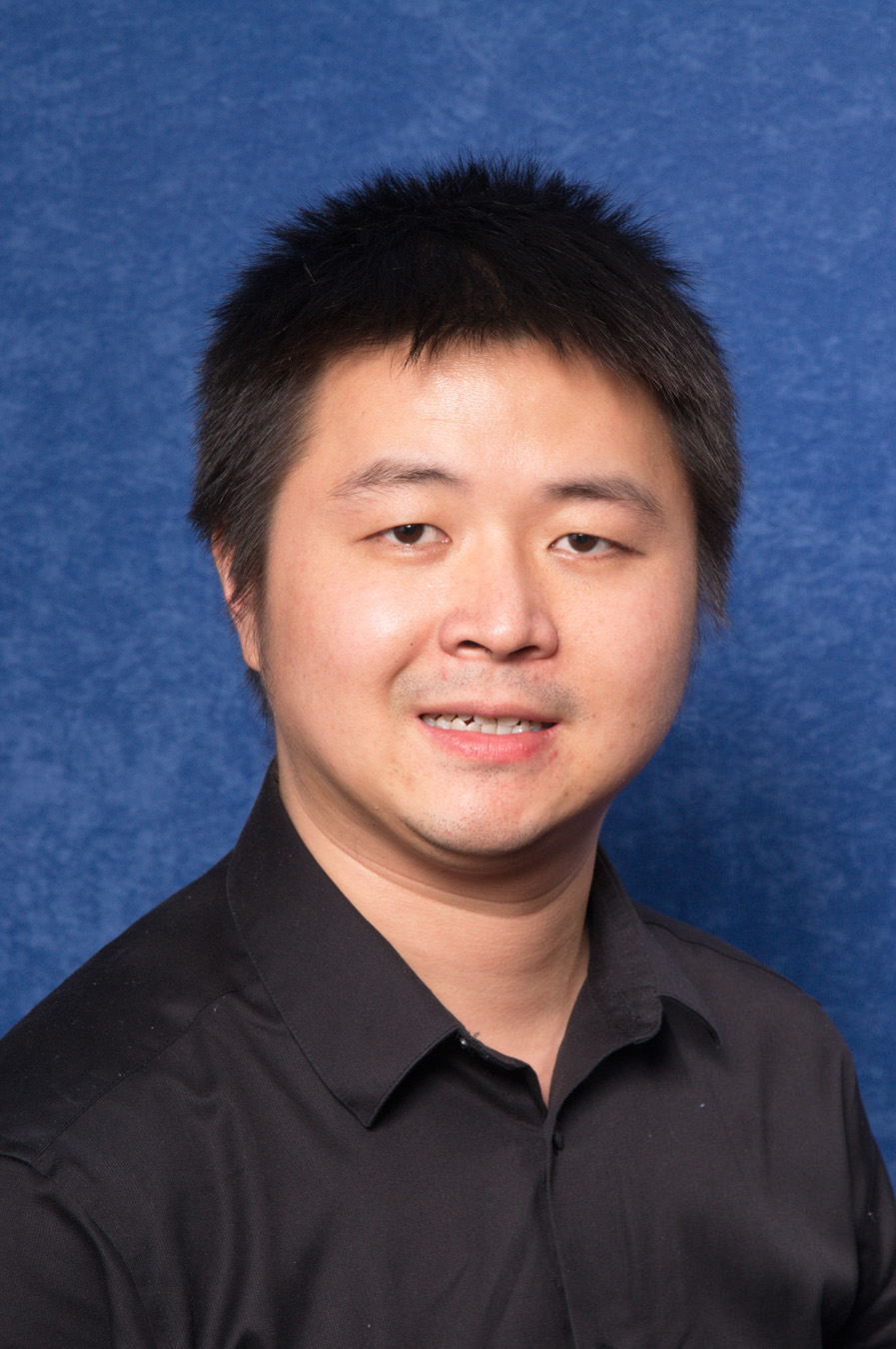}}]{Guo-Jun Qi}
Guo-Jun Qi (M14-SM18) is the Chief Scientist leading and overseeing an international R\&D team in the domain of multiple intelligent cloud services, including smart cities, visual computing service, medical intelligent service, and connected vehicle service at Futurewei, since August 2018. He was a faculty member in the Department of Computer Science and the director of MAchine Perception and LEarning (MAPLE) Lab at the University of Central Florida since August 2014. Prior to that, he was also a Research Staff Member at IBM T.J. Watson Research Center, Yorktown Heights, NY.
Dr. Qi has published over 150 papers in a broad range of venues.
Among them are the best student paper of ICDM 2014, ``the best ICDE 2013 paper" by IEEE Transactions on Knowledge and Data Engineering, as well as the best paper (finalist) of ACM Multimedia 2007 (2015).
\end{IEEEbiography}

\begin{IEEEbiography}[{\includegraphics[width=1in,height=1.25in,clip,keepaspectratio]{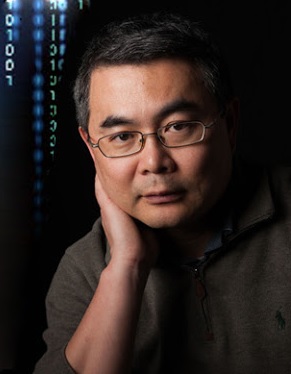}}]{Jiebo Luo}
Jiebo  Luo (S93-M96-SM99-F09)  joined  the  University  of  Rochester  in  Fall  2011  after  over  fifteen prolific years at Kodak Research Laboratories, where he was a Senior Principal Scientist leading research and advanced development. He has been involved in numerous  technical  conferences,  including  serving as the program co-chair of ACM Multimedia 2010,IEEE  CVPR  2012, ACM ICMR 2016,  and  IEEE  ICIP  2017.  He  has served on the editorial boards of the IEEE Transactions on Pattern Analysis and Machine Intelligence, IEEE  Transactions  on  Multimedia,  IEEE  Transactions  on  Circuits  and  Systems  for  Video  Technology, IEEE  Transactions  on  Big Data,   ACM  Transactions  on Intelligent Systems and Technology,  Pattern Recognition, Machine Vision and Applications, Knowledge and Information Systems, and Journal of Electronic Imaging. Dr. Luo is a Fellow of the SPIE, IAPR, IEEE, ACM, and AAAI.
\end{IEEEbiography}





\newpage
\newpage
\appendices

\renewcommand\thefigure{\thesection.\arabic{figure}}
\renewcommand\thetable{\thesection.\arabic{table}}

\section{Evaluations on Unsupervised Learning}\label{sec:ul_eval}

Unsupervised methods are often evaluated based on their performances on downstream tasks. In particular, the learned unsupervised representations can be used to perform classification tasks on benchmark datasets such CIFAR-10, ImageNet, Places and Pascal VOC.

For the sake of fair comparison, some efforts have been made on setting standard evaluation protocols on these datasets. In this subsection, we review such a protocol that has been widely adopted by many unsupervised methods.  Although not all approaches have been compared against this protocol due to the legacy reason, it has emerged in literature to allow a fair and direct comparison among many recent methods \cite{bojanowski2017unsupervised,gidaris2018unsupervised,oyallon2015deep,dosovitskiy2014discriminative,radford2015unsupervised,oyallon2017scaling,zhang2019aet}.


\subsection{Evaluation Protocol}

The evaluation of an unsupervised model often consists of two stages. The first stage is the unsupervised training of representations with only unlabeled examples.
In the second stage, a supervised classifier is trained on top of the learned representations to evaluate their performances on generalizability to a new classification task.

Take the evaluation protocol on the ImageNet dataset for example. The AlexNet is widely used as the backbone to learn the unsupervised representations, which consists of five convolutional layers and three fully connected layers (including a softmax layer with $1,000$-way outputs). There are several settings below adopted to test unsupervised models for classification tasks.

{\bf \noindent Nonlinear Classifiers} In this setting, the convolutional layers up to Conv4 and Conv5 are frozen after they are unsupervised trained in the first stage. All the layers above the Conv4 and Conv5 are supervised trained with the labeled data in the evaluation stage. In other words, the nonlinear classifiers are trained on the unsupervised representations for the evaluation purpose. For example, in the Conv4 setting, Conv5 and three fully connected layers are trained with the labeled examples, including the last $1000$-way output layer. Table~\ref{tab:nonlinear} shows the comparison between several unsupervised models. The fully supervised model (ImageNet Labels) and the random model are also included in the table to show the  upper and lower bounds on the classification performances, respectively.

\begin{table}
\caption{Top-1 accuracy with non-linear layers on ImageNet. AlexNet is used as backbone to train the unsupervised models. After unsupervised features are learned, nonlinear classifiers are trained on top of Conv4 and Conv5 layers with labeled examples to compare their performances. The fully supervised models and random models give upper and lower bounded performances.}\label{tab:nonlinear}
\centering
 \begin{tabular}{l|cc} \toprule
Method&Conv4 &Conv5\\ \midrule
ImageNet Labels \cite{bojanowski2017unsupervised}(Upper Bound)&59.7&59.7  \\
Random \cite{noroozi2017representation} (Lower Bound)&27.1 &12.0  \\ \midrule
Tracking \cite{wang2015unsupervised} &38.8&29.8 \\
Context \cite{doersch2015unsupervised} &45.6&30.4 \\
Colorization \cite{zhang2016colorful}&40.7&35.2 \\
Jigsaw Puzzles \cite{noroozi2016unsupervised}&45.3&34.6\\
BiGAN \cite{donahue2016adversarial}&41.9&32.2\\
NAT \cite{bojanowski2017unsupervised}&-&36.0\\
DeepCluster \cite{caron2018deep} &-&44.0\\
RotNet \cite{gidaris2018unsupervised}&50.0&43.8\\
AET \cite{zhang2019aet} &{53.2}&{47.0}\\
AVT \cite{qi2019avt}& \bf 54.2 & \bf 48.4\\\bottomrule
\end{tabular}
\end{table}

{\bf \noindent Linear Classifiers} A single fully connected layer can also be added on top of unsupervised representations to train a weak linear classifier. As shown in Table~\ref{tab:linear}, the linear layer is trained upon different convolutional layers of feature maps. The linear classifier can be trained very efficiently and the results show that a good trade off between training efficiency and test accuracy can be achieved with a linear classifier on top of a properly trained unsupervised representation.

\begin{table}
\scriptsize
\caption{Top-1 accuracy with linear layers on ImageNet. AlexNet is used as backbone to train the unsupervised models under comparison. A $1,000$-way linear classifier is trained upon various convolutional layers of feature maps that are spatially resized to have about $9,000$ elements. Fully supervised and random models are also reported to show the upper and the lower bounds of unsupervised model performances. Only a single crop is used and no dropout or local response normalization is used during testing, except the models denoted with * where ten crops are applied to compare results.}\label{tab:linear}
\centering
 \begin{tabular}{l|ccccc} \toprule
Method&Conv1 &Conv2&Conv3&Conv4&Conv5\\ \midrule
ImageNet Labels \cite{gidaris2018unsupervised}&19.3&36.3&44.2&48.3&50.5  \\
Random \cite{gidaris2018unsupervised} &11.6 &17.1&16.9&16.3&14.1  \\
Random rescaled \cite{krahenbuhl2015data}&17.5 &23.0&24.5&23.2&20.6  \\\midrule
Context \cite{doersch2015unsupervised} &16.2&23.3&30.2&31.7&29.6 \\
Context Encoders \cite{pathak2016context}&14.1&20.7&21.0&19.8&15.5 \\
Colorization\cite{zhang2016colorful}&12.5&24.5&30.4&31.5&30.3\\
Jigsaw Puzzles \cite{noroozi2016unsupervised}&18.2&28.8&34.0&33.9&27.1\\
BiGAN \cite{donahue2016adversarial}&17.7&24.5&31.0&29.9&28.0\\
Split-Brain \cite{zhang2017split}&17.7&29.3&35.4&35.2&32.8\\
Counting \cite{noroozi2017representation}&18.0&30.6&34.3&32.5&25.7\\
RotNet \cite{gidaris2018unsupervised}&18.8&31.7&38.7&38.2&36.5\\
Ins. Disc. \cite{wu2018unsupervised}&16.8&26.5&31.8&34.1&35.6\\
AET\cite{zhang2019aet}&{19.2}&{32.8}&{40.6}&{39.7}&{37.7}\\
AVT \cite{qi2019avt} & \bf 19.5 & \bf 33.6 & \bf 41.3 &\bf 40.3 &\bf  39.1 \\\midrule\midrule
DeepCluster* \cite{caron2018deep} &13.4&32.3&41.0&39.6&38.2\\
AET\cite{zhang2019aet}* &\bf {19.3}&\bf {35.4}&\bf {44.0}&\bf {43.6}&\bf {42.4}\\
\bottomrule
\end{tabular}\\
\end{table}

{\bf \noindent Cross-Dataset Tasks} Cross-dataset tasks are also performed to compare the generalizability of the unsupervised representations to the tasks on new datasets.

\begin{table}
\scriptsize
\caption{Top-1 accuracy on the Places dataset with linear layers. A $205$-way logistic regression classifier is trained on top of various layers of feature maps that are spatially resized to have about $9,000$ elements. All unsupervised features are pre-trained on the ImageNet dataset, which are frozen when training the logistic regression layer with Places labels. The unsupervised models are also compared with fully-supervised networks trained with Places Labels and ImageNet labels, along with random models. }\label{tab:places}
\centering
 \begin{tabular}{l|ccccc} \toprule
Method&Conv1 &Conv2&Conv3&Conv4&Conv5\\ \midrule
Places labels \cite{zhou2014learning}&22.1&35.1&40.2&43.3&44.6 \\
ImageNet labels&22.7&34.8&38.4&39.4&38.7\\
Random &15.7 &20.3&19.8&19.1&17.5  \\
Random rescaled \cite{krahenbuhl2015data}&21.4 &26.2&27.1&26.1&24.0  \\
\midrule
Context \cite{doersch2015unsupervised} &19.7&26.7&31.9&32.7&30.9 \\
Context Encoders \cite{pathak2016context}&18.2&23.2&23.4&21.9&18.4 \\
Colorization\cite{zhang2016colorful}&16.0&25.7&29.6&30.3&29.7\\
Jigsaw Puzzles \cite{noroozi2016unsupervised}&{23.0}&31.9&35.0&34.2&29.3\\
BiGAN \cite{donahue2016adversarial}&22.0&28.7&31.8&31.3&29.7\\
Split-Brain \cite{zhang2017split}&21.3&30.7&34.0&34.1&32.5\\
Counting \cite{zhang2017split}&{23.3}&{33.9}&{36.3}&{34.7}&29.6\\
RotNet \cite{gidaris2018unsupervised}&21.5&31.0&35.1&34.6&{33.7}\\
Ins. Dis. \cite{wu2018unsupervised}&18.8&24.3&31.9&34.5&33.6\\
AET \cite{zhang2019aet} &22.1&{32.9}&{37.1}&{36.2}&{34.7}\\
AVT \cite{qi2019avt} & 22.3 & 33.1 & 37.8 & 36.7 & 35.6\\\bottomrule
\end{tabular}
\end{table}

\begin{table}[t]
\scriptsize
\caption{Results on PASCAL VOC 2007 classification and detection tasks, and PASCAL VOC 2012 segmentation task. For classification, the convolutional features before Conv5 (fc6-8) or the whole model (all) are fine-tuned on PASCAL VOC dataset after they are unsupervised pretrained on ImageNet. For detection, multi-scale is used for training and a single scale for testing. The mean Average Precision (mAP) is reported for classification and detection tasks, while the mean Intersection over Union (mIoU) for the segmentation.}\label{tab:voc}
\centering
 \begin{tabular}{l|cccc} \toprule
Method&\multicolumn{2}{c}{Classification} &Detection&Segmentation\\ \midrule
Layers & fc6-8 & all & all & all\\\hline
ImageNet labels \cite{bojanowski2017unsupervised} & 78.9 & 79.9 & 56.8 & 48.0 \\
Random \cite{gidaris2018unsupervised} & - & 53.3 & 43.4 & 19.8 \\
Random rescaled \cite{krahenbuhl2015data} & 39.2 & 56.6 & 45.6 & 32.6 \\
Egomotion \cite{agrawal2015learning} & 31.0 & 54.2 & 43.9 & - \\
Context Encoders \cite{pathak2016context} & 34.6 & 56.5 & 44.5 & 29.7 \\
Tracking & 55.6 & 63.1 & 47.4 & - \\
Context \cite{doersch2015unsupervised} & 55.1 & 65.3 & 51.1 & - \\
Colorization \cite{zhang2016colorful} & 61.5 & 65.6 &  46.9 & 35.6 \\
BiGAN \cite{donahue2016adversarial} & 52.3 & 60.1 & 46.9 & 34.9 \\
Jigsaw Puzzles \cite{noroozi2016unsupervised} & - & 67.6 & 53.2 & 37.6 \\
NAT \cite{bojanowski2017unsupervised} & 56.7 & 65.3 & 49.4 & - \\
Split-Brain \cite{zhang2017split} & 63.0 & 67.1 & 46.7 & 36.0 \\
ColorProxy \cite{larsson2016learning} & - & 65.9 & - & - \\
Counting \cite{noroozi2017representation} & - & 67.7 & 51.4 & 36.6 \\
RotNet \cite{gidaris2018unsupervised} & 70.87 & 72.97 & 54.4 & 39.1 \\
AET \cite{zhang2019aet} & - & - & 57.4 & -\\ \bottomrule
\end{tabular}
\end{table}

\subsection{Results}

As shown in Table~\ref{tab:places}, unsupervised models have also been evaluated by pretraining on the ImageNet dataset. Then a single-layer logistic regression classifier is trained on top of different convolutional layers of feature maps with Places labels.
Table~\ref{tab:voc} shows the classification, object detection and semantic segmentations on PASCAL VOC, where the models are still based on AlexNet variants and pretrained on the ImageNet in an unsupervised fashion. Both results are compared against the fully supervised models trained with the Places labels and ImageNet labels, as well as the random networks.


\begin{table*}[h!]
\caption{Error rates on SVHN with a various number of labeled examples used to train different models. Note two versions of results have been reported on $\Pi$ model separately.}\label{tab:svhn}
\centering
 \begin{tabular}{l|cccc} \toprule
Method& 250 &500&1000&All Labels\\ \midrule
Supervised only \cite{tarvainen2017mean} & 27.77 $\pm$ 3.18 & 16.88 $\pm$ 1.30 & 12.32 $\pm$ 0.95 & 2.75 $\pm$ 0.10 \\
M1+M2 \cite{kingma2014semi} & - & - & 36.02 $\pm$ 0.10 & - \\
Improved GAN \cite{salimans2016improved} &-& 18.44$\pm$4.8 & 8.11$\pm1.3$ & - \\
ALI \cite{dumoulin2016adversarially} & - & - & 7.41 $\pm$ 0.65 & - \\
Localized GAN \cite{qi2018global} & - & 5.48 $\pm$ 0.29 & 4.79 $\pm$ 0.16 & - \\
$\Pi$ model \cite{laine2016temporal} & - & 6.65 $\pm$ 0.53 & 4.82 $\pm$ 0.17 & 2.54 $\pm$ 0.04 \\
$\Pi$ model \cite{tarvainen2017mean} & 9.69 $\pm$ 0.92 & 6.83 $\pm$ 0.60 & 4.95 $\pm$ 0.26 & 2.50 $\pm$ 0.07\\
Temporal Ensembling \cite{laine2016temporal} & - & 5.12 $\pm$ 0.13 & 4.42 $\pm$ 0.16 & 2.74 $\pm$ 0.06 \\
VAT + EntMin \cite{miyato2018virtual} & - & - & 3.86 & - \\
Mean Teacher \cite{tarvainen2017mean} & 4.35 $\pm$ 0.50 & 4.18 $\pm$ 0.27 & 3.95 $\pm$ 0.19 & 2.50 $\pm$ 0.05 \\\bottomrule
\end{tabular}
\end{table*}

\begin{table*}[h!]
\caption{Error rates on CIFAR-10 with a various number of labeled examples used to train different models. Note two versions of results have been reported on $\Pi$ model separately.}\label{tab:cifar10}
\centering
 \begin{tabular}{l|cccc} \toprule
Method& 1000 &2000&4000&All Labels\\ \midrule
Supervised only \cite{tarvainen2017mean} & 46.43 $\pm$ 1.21 & 33.94 $\pm$ 0.73 & 20.66 $\pm$ 0.57 & 5.82 $\pm$ 0.15 \\
Improved GAN \cite{salimans2016improved} &-& - & 18.63 $\pm$ 2.32 & - \\
ALI \cite{dumoulin2016adversarially} & - & - & 17.99 $\pm$ 1.62 & - \\
Localized GAN \cite{qi2018global} & 17.44 $\pm$ 0.25 & - & 14.23 $\pm$ 0.27 & - \\
$\Pi$ model \cite{laine2016temporal} & - & - & 12.36 $\pm$ 0.31 & 5.56 $\pm$ 0.10 \\
$\Pi$ model \cite{tarvainen2017mean} & 27.36 $\pm$ 1.20 & 18.02 $\pm$ 0.60 & 13.20 $\pm$ 0.27 & 6.06 $\pm$ 0.11\\
Temporal Ensembling \cite{laine2016temporal} & - & - & 12.16 $\pm$ 0.31 & 5.60 $\pm$ 0.10 \\
VAT + EntMin \cite{miyato2018virtual} & - & - & 10.55 & - \\
Mean Teacher \cite{tarvainen2017mean} & 21.55 $\pm$ 1.48 & 15.73 $\pm$ 0.31 & 12.31 $\pm$ 0.28 & 5.94 $\pm$ 0.15 \\\bottomrule
\end{tabular}
\end{table*}

\section{Evaluations on Semi-Supervised Learning}\label{sec:ssl_eval}

We will summarize some results by semi-supervised methods here. More evaluation protocols for comparing semi-supervised methods can be found in \cite{oliver2018realistic}.

\subsection{Datasets}
First we will introduce two datasets widely used in the evaluation.

{\noindent\bf CIFAR-10 Dasetset.}
The dataset \cite{krizhevsky2009learning} contains $50,000$ training images and $10,000$ test images on ten image categories. We train the semi-supervised LGAN model in experiments, where $100$ and $400$ labeled examples  are labeled per class and the remaining examples are left unlabeled.
The experiment results on this dataset are reported by averaging over ten runs.

{\noindent\bf SVHN Dataset.} The dataset \cite{netzer2011reading} contains $32\times 32$ street view house numbers that are roughly centered in images.  The training set and the test set contain $73,257$ and $26,032$ house numbers, respectively. In an experiment, $50$ and $1,00$ labeled examples per digit are used to train the model, and the remaining unlabeled examples are used as auxiliary data to train the model in semi-supervised fashion.

\subsection{Results}

Both CIFAR-10 and SVHN are often used to evaluate the performances of semi-supervised models by training
them with all unlabeled training images and a various amount of labeled examples. Then the error rate is reported on a separate test set.

For the sake of fair comparison, a 13-layer convolutional neural network is often adopted to train the models (See Table 5 of \cite{laine2016temporal}). For most of models, random translations and horizontal flips are applied as data augmentations of input images. There are also two forms of noises used in many models (e.g., $\Pi$ model, mean teacher, Temporal ensembling): Gaussian noises are applied to the input layers while the random dropout is applied within the networks.

Table~\ref{tab:cifar10} and Table~\ref{tab:svhn} compare the results on CIFAR-10 and SVHN datasets respectively, from which we can see that the class of teach-student models turn out to outperform other methods well on both datasets. In particular, VAT has achieved the most outstanding performances among these compared models.

\section{Chart of Unsupervised and Semi-Supervised Learning}

\begin{figure*}[h]
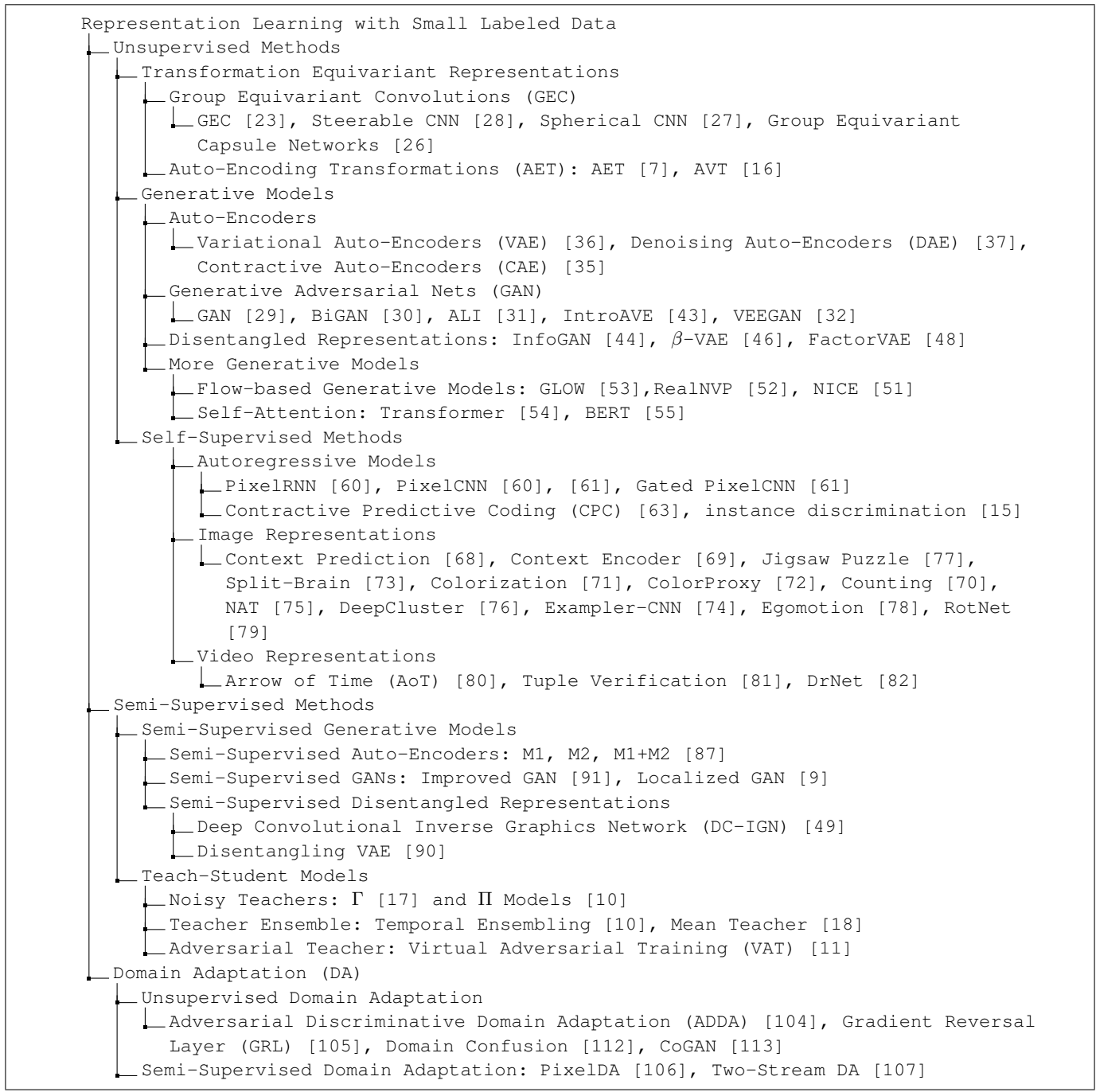

\centering
\framebox[\textwidth]{%
\small\small
\begin{minipage}[c]{0.9\textwidth}
\fontfamily{times}
\dirtree{%
.1 Representation Learning with Small Labeled Data.
.2 Unsupervised Methods.
.3 Transformation Equivariant Representations.
.4 Group Equivariant Convolutions (GEC).
.5 GEC \cite{cohen2016group}, Steerable CNN \cite{cohen2016steerable}, Spherical CNN \cite{cohen2018spherical}, Group Equivariant Capsule Networks \cite{lenssen2018group}.
.4 Auto-Encoding Transformations (AET): AET \cite{zhang2019aet}, AVT \cite{qi2019avt}.
.3 Generative Models.
.4 Auto-Encoders.
.5 Variational Auto-Encoders (VAE) \cite{kingma2013auto}, Denoising Auto-Encoders (DAE) \cite{vincent2008extracting}, Contractive Auto-Encoders (CAE) \cite{rifai2011contractive}.
.4 Generative Adversarial Nets (GAN).
.5 GAN \cite{goodfellow2014generative}, BiGAN \cite{donahue2016adversarial}, ALI \cite{dumoulin2016adversarially}, IntroAVE \cite{huang2018introvae}, VEEGAN \cite{srivastava2017veegan}.
.4 Disentangled Representations: InfoGAN \cite{chen2016infogan}, $\beta$-VAE \cite{higgins2016beta}, FactorVAE \cite{kim2018disentangling}.
.4 More Generative Models.
.5 Flow-based Generative Models: GLOW \cite{kingma2018glow},RealNVP \cite{dinh2016density}, NICE \cite{dinh2014nice}.
.5 Self-Attention: Transformer \cite{vaswani2017attention}, BERT \cite{devlin2018bert}.
.3 Self-Supervised Methods.
.5 Autoregressive Models.
.6 PixelRNN \cite{oord2016pixel}, PixelCNN \cite{oord2016pixel,van2016conditional}, Gated PixelCNN \cite{van2016conditional}.
.6 Contractive Predictive Coding (CPC) \cite{oord2018representation}, instance discrimination \cite{wu2018unsupervised}.
.5 Image Representations.
.6 Context Prediction \cite{doersch2015unsupervised}, Context Encoder \cite{pathak2016context}, Jigsaw Puzzle \cite{noroozi2017representation},  Split-Brain \cite{zhang2017split}, Colorization \cite{zhang2016colorful}, ColorProxy \cite{larsson2016learning}, Counting \cite{noroozi2016unsupervised}, NAT \cite{bojanowski2017unsupervised}, DeepCluster \cite{caron2018deep}, Exampler-CNN \cite{dosovitskiy2014discriminative}, Egomotion \cite{agrawal2015learning}, RotNet \cite{gidaris2018unsupervised}.
.5 Video Representations.
.6 Arrow of Time (AoT) \cite{wei2018learning}, Tuple Verification \cite{misra2016shuffle}, DrNet \cite{denton2017unsupervised}.
.2 Semi-Supervised Methods.
.3 Semi-Supervised Generative Models.
.4 Semi-Supervised Auto-Encoders: M1, M2, M1+M2 \cite{kingma2014semi}.
.4 Semi-Supervised GANs: Improved GAN \cite{salimans2016improved}, Localized GAN \cite{qi2018global}.
.4 Semi-Supervised Disentangled Representations.
.5 Deep Convolutional Inverse Graphics Network (DC-IGN) \cite{kulkarni2015deep}.
.5 Disentangling VAE \cite{narayanaswamy2017learning}.
.3 Teach-Student Models.
.4 Noisy Teachers: $\Gamma$ \cite{rasmus2015semi} and $\Pi$ Models \cite{laine2016temporal}.
.4 Teacher Ensemble: Temporal Ensembling \cite{laine2016temporal}, Mean Teacher \cite{tarvainen2017mean}.
.4 Adversarial Teacher: Virtual Adversarial Training (VAT) \cite{miyato2018virtual}.
.2 Domain Adaptation (DA).
.3 Unsupervised Domain Adaptation.
.4 Adversarial Discriminative Domain Adaptation (ADDA) \cite{tzeng2017adversarial}, Gradient Reversal Layer (GRL) \cite{ganin2016domain}, Domain Confusion \cite{tzeng2015simultaneous}, CoGAN \cite{liu2016coupled}.
.3 Semi-Supervised Domain Adaptation: PixelDA \cite{bousmalis2017unsupervised}, Two-Stream DA \cite{rozantsev2018beyond}.
}
\end{minipage}
}
\caption{The chart presents the categorization of unsupervised and semi-supervised methods. The methods are organized by where they are reviewed in this survey.}
\label{fig:small_data}
\end{figure*}

Finally, we present a chart in Figure~\ref{fig:small_data} showing the categorization of unsupervised and semi-supervised methods.  This is for readers' convenience to look up the relevant methods reviewed in this survey.

\end{document}